\documentclass[letterpaper]{article} 
\usepackage{aaai23}  
\usepackage{times}  
\usepackage{helvet}  
\usepackage{courier}  
\usepackage[hyphens]{url}  
\usepackage{graphicx} 
\usepackage{natbib}  
\usepackage{caption} 
\usepackage{algorithm}
\usepackage{algorithmic}
\usepackage{caption}
\usepackage{subcaption}
\usepackage{multirow}
\usepackage{xspace}
\usepackage{makecell}
\usepackage{verbatim}

\usepackage[utf8]{inputenc} 
\usepackage[T1]{fontenc}    
\usepackage{url}            
\usepackage{booktabs}       
\usepackage{amsfonts}       
\usepackage{nicefrac}       
\usepackage{microtype}      
\usepackage{bm}
\usepackage{enumitem}
\usepackage{graphicx}
\usepackage{amsmath}
\usepackage{caption}
\usepackage{algorithm}
\usepackage{algorithmic}
\usepackage[dvipsnames]{xcolor}

\usepackage{enumitem}

\DeclareMathOperator*{\argmax}{arg\,max}
\usepackage[colorlinks=true,bookmarksopen,pdfsubject={algorithms},linkcolor={MidnightBlue}, anchorcolor={black}, citecolor={MidnightBlue}, filecolor={magenta}, menucolor={black}, pagecolor={red},backref=none,urlcolor={MidnightBlue}]{hyperref}
\usepackage{array}
\usepackage{multirow}
\usepackage{wrapfig,lipsum,booktabs}
\usepackage{newfloat}
\usepackage{listings}
\floatstyle{ruled}
\newfloat{listing}{tb}{lst}{}
\floatname{listing}{Listing}
\title{Accelerating Inverse Learning via Intelligent Localization with \\ Exploratory Sampling}
\author {
    Jiaxin Zhang\textsuperscript{\rm 1},
    Sirui Bi\textsuperscript{\rm 2},
    Victor Fung\textsuperscript{\rm 3}
}
\affiliations {
    \textsuperscript{\rm 1} Intuit AI Research \\ 
    \textsuperscript{\rm 2} Walmart Global Tech \\ 
    \textsuperscript{\rm 3} Georgia Institute of Technology\\
    jiaxin\_zhang@intuit.com, sirui.bi@walmart.com, victorfung@gatech.edu \\
}

\usepackage{bibentry}

\begin{document}

\maketitle

\begin{abstract}
In the scope of ``AI for Science'', solving inverse problems is a longstanding challenge in materials and drug discovery, where the goal is to determine the hidden structures given a set of desirable properties. Deep generative models are recently proposed to solve inverse problems, but these currently use expensive forward operators and struggle in precisely localizing the exact solutions and fully exploring the parameter spaces without missing solutions. In this work, we propose a novel approach (called {iPage}) to accelerate the inverse learning process by leveraging probabilistic inference from deep invertible models and deterministic optimization via fast gradient descent. Given a target property, the learned invertible model provides a posterior over the parameter space; we identify these posterior samples as an intelligent prior initialization which enables us to narrow down the search space. We then perform gradient descent to calibrate the inverse solutions within a local region. Meanwhile, a space-filling sampling is imposed on the latent space to better explore and capture all possible solutions. We evaluate our approach on three benchmark tasks and two created datasets with real-world applications from quantum chemistry and additive manufacturing, and find our method achieves superior performance compared to several state-of-the-art baseline methods. The iPage code is available at \url{https://github.com/jxzhangjhu/MatDesINNe}.
\end{abstract}

\section{Introduction}
\label{intro}

A fundamental problem in materials and drug discovery is to find novel structures (e.g., molecules or crystals) with desirable properties. One typical approach is to search in the chemical space based on a specific property prediction. Inverse design provides a promising way for this problem by inverting this paradigm by starting with the desired functionality and searching for an ideal molecular structure \citep{sanchez2018inverse, yao2021inverse}, as opposed to the direct approach that maps from existing molecules in chemical space to the properties. Mathematically speaking, inverse design tries to solve a nonlinear inverse problem, which remains a significant challenge in natural sciences and mathematics, and also plays a critical role in safe decision-making with uncertainty. Typically, the approach is to develop a mathematical operator or physical model $\Omega$ on how measured observations (e.g., properties) $\mathbf{y} \in \mathbb{R}^{M}$ arise from the input hidden parameters (e.g., chemical space) $\mathbf{x} \in \mathbb{R}^D$ and such mapping $\mathbf{y}=\Omega(\mathbf{x})$ represents the {\em forward process}. The opposite direction, the {\em inverse process} $\mathbf{x} = \Omega^{-1}(\mathbf{y})$, involves the inference of the hidden parameters from measurements. However, the inverse process is ill-posed with a one-to-many mapping such that finding $\Omega^{-1}$ becomes intractable. 

Unfortunately, inverse design in scientific exploration differs from conventional inverse problems in that it poses several unique additional challenges. First, \emph{the forward operator $\Omega$ is not explicitly known}. In many cases, the forward operator is modeled by first-principles calculations or large-scale complex simulations \cite{lavin2021simulation}, including molecular dynamics and density functional theory \cite{liu2022machine}. This challenge makes inverse design difficult to leverage recent advances in solving inverse problems, such as MRI reconstruction \citep{wang2020deep}, implanting on images through generative models with large datasets \citep{asim2020invertible}. Second, \emph{the search space $\mathbf{x} \in \mathbb{R}^D$ is often huge}. For example, small drug-like molecules have been estimated to contain between 10\textsuperscript{23} to 10\textsuperscript{60} unique cases 
\citep{ertl2009estimation}, while solid materials have an even larger space. This challenge results in obvious obstacles for using global search via Bayesian optimization or using Bayesian inference via Markov Chain Monte Carlo (MCMC) since either method is prohibitively slow for high-dimensional inverse problems \cite{zhang2019learning}. Third, \emph{multimodal solutions have a one-to-many mapping issue}. In other words, there are multiple solutions that match the desirable property. This issue leads to difficulties in pursuing all possible solutions through a gradient-based optimization, which converges a single deterministic solution and is easily trapped into local minima \cite{zhang2021enabling}. Probabilistic inference also has limitations in approximating complex posteriors, which may cause the simple point estimates (e.g., maximum a posterior (MAP)) to have misleading solutions \citep{sun2020deep}. This work aims at addressing these challenges by leveraging advantages from probabilistic inference and deterministic optimization to accelerate solving of generic inverse problems. The key insight is to first learn an approximate posterior distribution by training an invertible neural network (INN) given paired datasets $\mathcal{D} = \{\mathbf{x}_i, \mathbf{y}_i \}_{i=1}^m$ and then perform a local search via optimization by starting with these posterior samples as good initialization (called ``intelligent priors''). 

Specifically, we propose a dynamic bi-directional training scheme to obtain a backward model and a forward model simultaneously. The backward model is used to generate posterior samples $\mathbf{x}^*$ given target properties $\mathbf{y}^*$. These posterior samples $\mathbf{x}^*$ as intelligent priors significantly narrow down the search space from the entire domain to a local domain. The forward model serves as a surrogate of the forward operator $\Omega$ and provides an accurate gradient estimate with respect to the design space $\mathbf{x} \in \mathbb{R}^D$ via automatic differentiation. This enables us to localize all possible solutions simultaneously by conducting an efficient gradient-based optimization starting from the intelligent priors $\mathbf{x}^*$. Unlike direct global search with random initialization, our method significantly accelerates inverse learning by conducting an efficient local search via gradient descent on low-dimensional design space. Compared with probabilistic inference via unsupervised generative models, our inverse solutions are closer to the ground truth since a supervised localization scheme is performed on the posterior samples. Our method is also applicable to high-dimensional problems with relatively small datasets given an expensive forward operator $\Omega$. More importantly, we propose an exploratory sampling strategy with an enhanced space-filling capability to better explore and capture all possible solutions in the design space. To the best of our knowledge, this is the \emph{first work} to investigate space-filling sampling on the latent variable of the INN model to improve sampling exploration with variance reduction. As a result, we achieve superior performance in re-simulation accuracy, space exploration, and solution diversity through multiple artificial benchmarks. We also curate two real-world datasets from quantum chemistry and additive manufacturing and create a set of physically meaningful tasks and metrics for the problem of inverse learning. We find our method achieves superior performance compared to several state-of-the-art baseline methods.

\section{Related Work}
{\bf Bayesian and Variational Approaches.}
From the inference perspective, solving inverse problems can be achieved by estimating the full posterior distributions of the parameters conditioned on a target property. Bayesian methods, such as approximate Bayesian computing \citep{yang2018predictive}, are ideal choices to model the conditional posterior but this idea still encounters various computational challenges in high-dimensional cases \cite{zhang2018effect,zhang2018quantification}. An alternative choice are variational approaches, e.g., conditional GANs \citep{wang2018high} and conditional VAEs \citep{sohn2015learning}, which enable the efficient approximation of the true posterior by learning the transformation between latent variables and parameter variables. However, the direct application of both conditional generative models for inverse problems is challenging because a large dataset is often required \citep{tonolini2020variational}.

\vspace{0.2cm}
\noindent {\bf Deep Generative Models.} 
Many recent efforts have been made on solving inverse problems via deep generative models \citep{asim2020invertible, whang2021composing, whang2021solving, daras2021intermediate, sun2020deep, kothari2021trumpets,song2021solving}. For example, \citet{asim2020invertible} focuses on producing a point estimate motivated by the MAP formulation and \cite{whang2021composing} aims at studying the full distributional recovery via variational inference. A follow-up study from \cite{whang2021solving} is to study image inverse problems with a normalizing flow prior. For MRI or implanting on images, strong baseline methods exist that benefit from explicit forward operator \citep{sun2020deep, asim2020invertible, kothari2021trumpets}. We do not expect any benefit from using our method here. Instead, our focus is on the inverse design perspective where paired data $\mathcal{D} = \{\mathbf{x}_i, \mathbf{y}_i \}_{i=1}^m$ is limited since the forward operator is not explicitly known and is often computationally intensive. 

\vspace{0.2cm}
 \noindent{\bf Invertible Models.} 
Flow-based models \citep{rezende2015variational,dinh2016density,kingma2018glow,grathwohl2018ffjord,wu2020stochastic,nielsen2020survae}, may offer a promising direction to infer the posterior by training on invertible architectures. Some recent studies have leveraged this unique property of invertible models to address several challenges in solving inverse problems \citep{ardizzone2018analyzing, kruse2021benchmarking}. However, these existing invertible model approaches suffer from limitations \citep{ren2020benchmarking} in fully exploring the parameter space, leading to missed potential solutions, and often fail to precisely localize the optimal solutions due to noisy solutions and inductive errors, specifically in materials design problems \cite{fung2021inverse,fung2022atomic}.  

\vspace{0.2cm}
\noindent{\bf Surrogate-based Optimization.}
Another approach is to build a neural network surrogate and then conduct surrogate-based optimization via gradient descent. This is common in scientific and engineering applications \citep{forrester2009recent, gomez2018automatic, white2019multiscale}. The essential challenge is that the forward model is often time-consuming so a faster surrogate enables an intractable search. A recent study in the scope of surrogate-based optimization is the neural-adjoint (NA) method \citep{ren2020benchmarking} which directly searches the global space via gradient descent starting from random initialization, such that a large number of interactions are required to converge and its solutions are easily trapped in the local minima \citep{deng2021neural}. Although the neural-adjoint (NA) method boosts the performance by down-selecting the top solutions from multiple starts, the computational cost is significantly high, specifically for high-dimensional problems. 

\section{Methodology}
\subsection{Augmented Inverse Learning Formulation}
\begin{figure}[!h]
  \centering
  \includegraphics[width=0.95\linewidth]{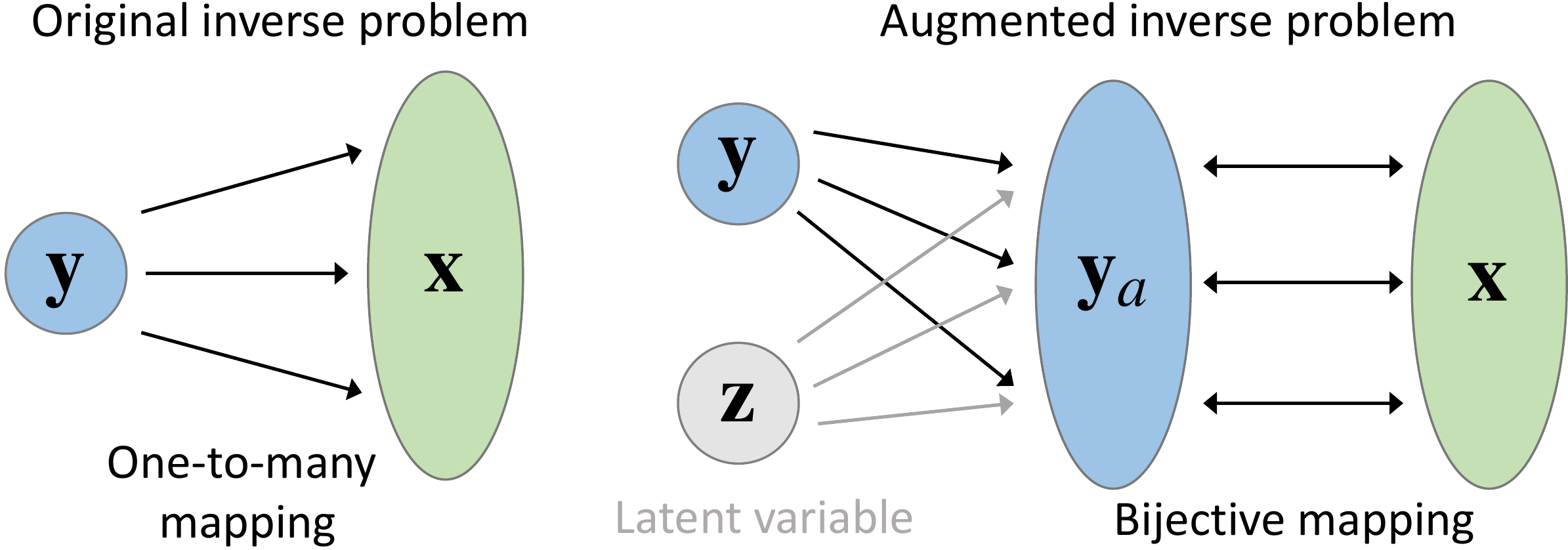}
  \caption{Original inverse problem often encounters the ill-posed issue due to one-to-many mappings. An augmented inverse problem is formulated based on bijective mapping with additional latent variable $\mathbf{z}$.}
\label{fig:eq_bijectiv}
  \vspace{-6mm}
\end{figure}

In natural sciences, a mathematical or physical model is often developed to describe how measured observations $\mathbf{y} \in \mathbb{R}^{M}$ arise from the hidden parameters $\mathbf{x} \in \mathbb{R}^D$, to yield such a mapping $\mathbf{y}=\Omega(\mathbf{x})$. To completely capture all possible inverse solutions given observed measurements, a proper inverse model should enable the estimation of the full posterior distribution $p(\mathbf{x}|\mathbf{y})$ of hidden parameters $\mathbf{x}$ conditioned on an observation $\mathbf{y}$. One promising approach is to approximate $p(\mathbf{x}|\mathbf{y})$ with a tractable probabilistic model $\hat{p}(\mathbf{x}|\mathbf{y})$ by leveraging the advantage of the flexibility to generate paired training data $\left\{(\mathbf{x}_i, \mathbf{y}_i)_{i=1}^N \right\}$ from the well-understood forward process $\mathbf{y}_i = \Omega(\mathbf{x}_i)$. Invertible neural networks (INNs) \citep{ rezende2015variational,dinh2016density,ardizzone2018analyzing} can be trained in the forward process and then used in the invertible mode to sample from $p(\mathbf{x}|\mathbf{y})$ for any specific $\mathbf{y}$. This is achieved by adding a latent variable $\mathbf{z} \in \mathbb{R}^K$, which encodes the inherent information loss in the forward process. In other words, the latent variable $\mathbf{z}$ drawn from a Gaussian distribution $p(\mathbf{z}) = \mathcal{N}(0,I_K)$ is able to encode the intrinsic information about $\mathbf{x}$ that is {\em not} contained in $\mathbf{y}$. To this end, an augmented inverse problem is formulated based on such a bijective mapping (Fig. \ref{fig:eq_bijectiv}):
\begin{equation}
    \mathbf{x} = h(\mathbf{y}_a; \phi) = h(\mathbf{y},\mathbf{z}; \phi), \quad \mathbf{z} \sim p(\mathbf{z}) \label{eq:bijective}
\end{equation}
where $h$ is a function of $\mathbf{y}$ and $\mathbf{z}$, parametrized by an INN with parameters $\phi$. Forward training optimizes the mapping $\mathbf{x} \rightarrow \mathbf{y}_a = [\mathbf{y}, \mathbf{z}]$ and implicitly determines the inverse mapping $\mathbf{x} = h(\mathbf{y},\mathbf{z})$. In the context of INNs, the posterior distribution $p(\mathbf{x}|\mathbf{y})$ is represented by the deterministic function $\mathbf{x}=h(\mathbf{y},\mathbf{z})$ that transforms the known probability distribution $p(\mathbf{z})$ to parameter $\mathbf{x}$-space, conditional on measurements $\mathbf{y}$. Thus, given a chosen observation $\mathbf{y}^*$ with the learned $h$, we can obtain the posterior samples $\mathbf{x}_k$ which follows the posterior distribution $p(\mathbf{x} | \mathbf{y}^*)$ via a transformation $\mathbf{x}_k = h(\mathbf{y}^*,\mathbf{z}_k)$ with prior samples drawn from $\mathbf{z}_k \sim p(\mathbf{z})$.

The invertible architecture simultaneously learns the model $h(\mathbf{y},\mathbf{z}; \phi)$ of the inverse process jointly with a model $f(\mathbf{x};\phi)$ which approximates the true forward process $\Omega(\mathbf{x})$:
\begin{equation}
    [\mathbf{y},\mathbf{z}] = f(\mathbf{x};\phi) = [f_{\mathbf{y}}(\mathbf{x};\phi), f_{\mathbf{z}}(\mathbf{x};\phi)] = h^{-1}(\mathbf{x};\phi)
\end{equation}
where $f_{\mathbf{y}}(\mathbf{x};\phi) \approx \Omega(\mathbf{x})$, model $f$ and $h$ share the same parameters $\phi$ in a single invertible neural network. Therefore, our approximated posterior model $\hat{p}(\mathbf{x}|\mathbf{y})$ is built into the invertible neural network representation
\begin{equation}
    \hat{p}(\mathbf{x} = h(\mathbf{y},\mathbf{z}; \phi) | \mathbf{y}) = p(\mathbf{z})/ \left| \bm J_{\mathbf{x}} \right| 
\end{equation}
where the Jacobian $\bm J_{\mathbf{x}}$ can be efficiently computed by using neural spline flows \citep{durkan2019neural}. 

\subsection{Dynamical Bi-directional Training}
To optimize the loss more effectively, we perform a dynamic bi-directional training scheme by accumulating gradients from both forward and backward directions before updating the parameters, using an adaptive update strategy for the forward and backward loss weights $\lambda$. Specifically, the INN training is performed by minimizing the total loss:
\begin{equation}
    \mathcal{L}_{\textit{total}} = \lambda_x \mathcal{L}_{x} +  \lambda_y \mathcal{L}_{y} + \lambda_z \mathcal{L}_{z} \label{eq:total_loss}
\end{equation}
where $\mathcal{L}_{y}$ is a forward supervised loss that matches the neural network prediction $f_{\mathbf{y}}(\mathbf{x}_k;\phi)$ to the true observation via known forward simulation $\mathbf{y}_k = \Omega(\mathbf{x}_k)$.
\begin{equation}
    \mathcal{L}_{y} = \sum_{k=1}^N || f_{\mathbf{y}}(\mathbf{x}_k;\phi) - \mathbf{y}_k  ||^2. \label{eq:loss_y}
\end{equation}
$\mathcal{L}_z$ is an unsupervised loss for the latent variable, which penalizes deviations between the joint distribution $\hat{p}(\mathbf{y} = f_{\mathbf{y}}(\mathbf{x}),\mathbf{z}=f_{\mathbf{z}}(\mathbf{x}))$ and the product of the latent distribution $p(\mathbf{z})$ and the marginal distributions of $p(\mathbf{y}=\Omega(\mathbf{x}))$:
\begin{equation}
    \mathcal{L}_{z} = \textup{MMD}\left\{f(\mathbf{x}_k; \phi); p(\mathbf{y}) p(\mathbf{z})) \right\} \label{eq:loss_z}
\end{equation}
where MMD refers to the Maximum Mean Discrepancy \citep{gretton2012kernel,gretton2012optimal}, a kernel-based approach that only requires samples from each probability distribution to be compared. Practically, $\mathcal{L}_{z}$ enforces $\mathbf{z}$ follow the desired Gaussian distribution $p(\mathbf{z})$, and ensures $\mathbf{z}$ and $\mathbf{y}$ are independent without sharing the same information. 

$\mathcal{L}_{\mathbf{x}}$ is an unsupervised loss, which is implemented by MMD and used to penalize the mismatch between the distribution of backward predictions and the prior data distribution $p(\mathbf{x})$ if it is known,
\begin{equation}
   \mathcal{L}_{\mathbf{x}} = \textup{MMD}\left\{f^{-1}(\mathbf{y}_k, \mathbf{z}_k; \phi), p(\mathbf{x}) \right\}  \label{eq:loss_x}
\end{equation}
where $\mathcal{L}_{\mathbf{x}}$ aims to improve convergence and does not interfere with optimization. Theoretically, if $\mathcal{L}_{\mathbf{y}}$ and $\mathcal{L}_{\mathbf{z}}$ has converged to zero, and $\mathcal{L}_{\mathbf{x}}$ is guaranteed to be zero so that the samples drawn from Eq.~\eqref{eq:bijective} will follow the true posterior $p(\mathbf{x}|\mathbf{y}^*)$ for any observation $\mathbf{y}^*$. Therefore, a point estimate from the true posterior will lead to an exact inverse solution. However, practically, due to a finite training time, there is always a difference between the $\mathcal{L}_{\textup{total}}$ and zero loss, as well as a residual dependency between $\mathbf{y}$ and $\mathbf{z}$. This causes a mismatch between the approximated posterior $\hat{p}(\mathbf{x}|\mathbf{y})$ and the true posterior $p(\mathbf{x}|\mathbf{y})$. 

Our objective is to minimize the mismatch by optimization with a good initialization. Assuming $n_{t}$ training epochs are used, we set an initial large weight for the supervised loss $\lambda_{\mathbf{y}}^i \rightarrow N_{\ell}, i =1,...,n_t/2, \ N_{\ell} \gg 1$ to seek an accurate regression model $f_{\mathbf{y}}(\mathbf{x};\phi)$ and then perform an adaptive decay when $i = n_t/2,..,n_t$ and ensure $\lambda_{\mathbf{y}}^{n_t} \rightarrow 0$ at the end of training. The model with minimal $\ell_2$ loss $f_{\mathbf{y}}(\mathbf{x};\phi^*)$ is saved for prediction and gradient estimation. Meanwhile, the weights of unsupervised loss $\lambda_{\mathbf{x}}$ and $\lambda_{\mathbf{z}}$ are set by $\lambda_{\mathbf{x}}^i \rightarrow 0$ and $\lambda_{\mathbf{z}}^i \rightarrow 0$ when $i=1,...,n_t/2$ and are then adaptively increased until $\lambda_{\mathbf{x}}^{n_t} \rightarrow N_{\ell}$ and $\lambda_{\mathbf{z}}^{n_t} \rightarrow N_{\ell}$ where we minimize the residual dependency between $\mathbf{y}$ and $\mathbf{z}$ to approximate the true posterior  $p(\mathbf{x}|\mathbf{y})$. To do so, the backward MMD loss is minimized such that the learned posterior will be closer to the true posterior. 
\begin{figure*}[h!]
\vspace{-0.2cm}
    \centering
    \includegraphics[width=0.99\textwidth]{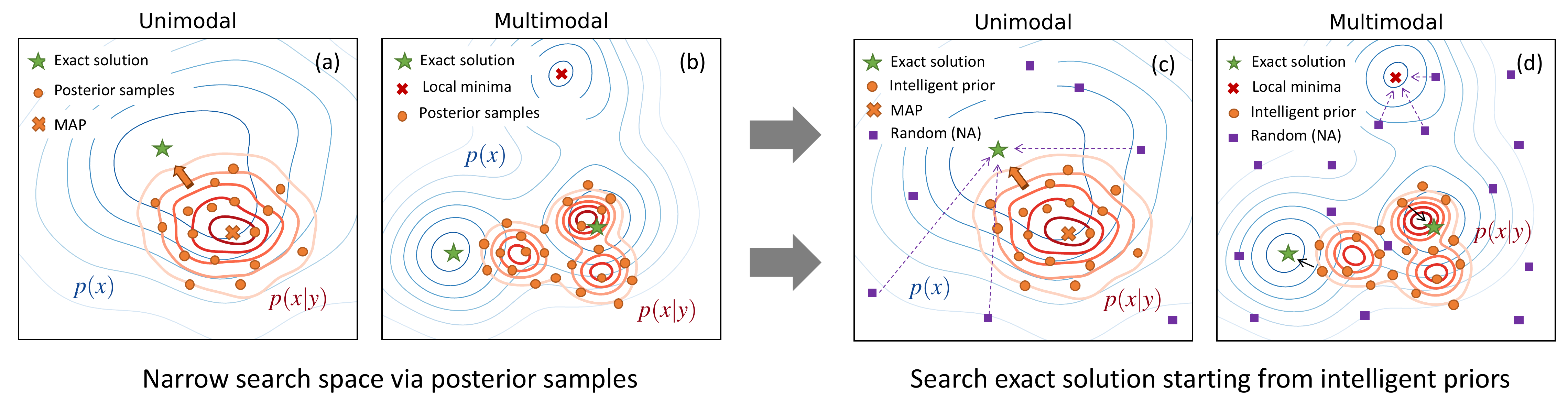}
    \vspace{-0.2cm}
    \caption{{Localizing inverse solutions from intelligent priors (posterior samples)}. The approximated posterior and its MAP estimator both deviate from the exact solution but successfully narrow down the search space. Our objective is to localize the exact solution by leveraging these posterior samples as intelligent initialization such that the process can be accelerated.}
    \label{fig:localize}
\end{figure*}

\subsection{Localization from Posterior Samples}
After finishing the dynamic bi-directional training, a set of posterior samples can be drawn from the approximated posterior distribution $\hat{p}(\mathbf{x}|\mathbf{y})$, as the orange dots shown in Fig. \ref{fig:localize} (left). Compared with the prior distribution $p(\mathbf{x})$, these posterior samples drawn from ${p}(\mathbf{x}|\mathbf{y})$ in either are able to reduce the search space with a smaller gap from the exact solution, which fits both unimodal and multimodal scenarios. Instead of global random search used by surrogate-based optimization, we localize the solutions starting from these posterior samples which can be seen as \emph{intelligent priors} (see Fig. \ref{fig:localize} (right)) and the smaller gap can be quickly filled by gradient descent with few steps. The local search via optimization would significantly accelerate the localization process and decrease the risk of local minima in global random search. This novel idea mainly consists of three steps: 

\begin{itemize}[leftmargin=10pt]
    \item {\bf Step 1 (Prior Exploration)}: Given a specific target $\hat{\mathbf{y}}$, repeat for a latent sample $\left\{\mathbf{z}_i \sim p(\mathbf{z}) \right\}_{i=1}^m$ to obtain a posterior sample $\left\{\hat{\mathbf{x}}_i \sim \hat{p}({\mathbf{x}} | \hat{\mathbf{y}}) \right\}_{i=1}^m$ which can be interpreted as a prior exploration of the solution space. Compared to the samples $\mathbf{x}_i$ directly drawn from the prior distribution $p(\mathbf{x})$, these posterior samples $\hat{\mathbf{x}}_i$ serve as good initialization, significantly shorten the distance to the exact inverse solution, as explained in Fig. \ref{fig:localize} (a) and (b). 
    \item {\bf Step 2 (Gradient Estimation)}: Extract the saved regression model $\hat{f}_{\mathbf{y}}(\mathbf{x};{\phi}^*)$ where the neural network parameters ${\phi}^*$ are fixed, and evaluate the model only by changing the input $\mathbf{x}$ to the network. The gradient at the current input $\hat{\mathbf{x}}_i$ can be defined as
    \begin{equation}
        \mathbf{g}_i = \frac{\partial\mathcal{L}(\hat{f}_{\mathbf{y}}(\hat{\mathbf{x}}_i;{\phi}^*), \hat{\mathbf{y}})}{\partial \mathbf{x}} \bigg|_{\mathbf{x} =\hat{\mathbf{x}}_i} \quad \hat{\mathbf{x}}_i \sim \hat{p}({\mathbf{x}} | \hat{\mathbf{y}}) \label{eq:gradient}
    \end{equation}
    where $\mathcal{L}$ is the $\ell_2$ loss and the gradient $\mathbf{g}_i$ can be efficiently computed by automatic differentiation. 
    \item {\bf Step 3 (Solution Localization)}: Precisely localize the posterior samples drawn from $\hat{p}(\mathbf{x} | \mathbf{y})$ to exact inverse solutions via gradient descent $\hat{\mathbf{x}}_i^{k+1} = \hat{\mathbf{x}}_i^{k} -\gamma ~ \mathbf{g}_i^{k}$,
    where $\gamma$ is the learning rate. We use Adam as the optimizer to adaptively update the solution. Compared with the generic random search in the entire space, our local search with intelligent priors is much more efficient and the bad (local) minima issue is naturally mitigated. 
\end{itemize}

\subsection{Space-filling Sampling on Latent Space}
\begin{figure}[t]
\centering
    \vspace{-0.2cm}
    \includegraphics[width=0.47\textwidth]{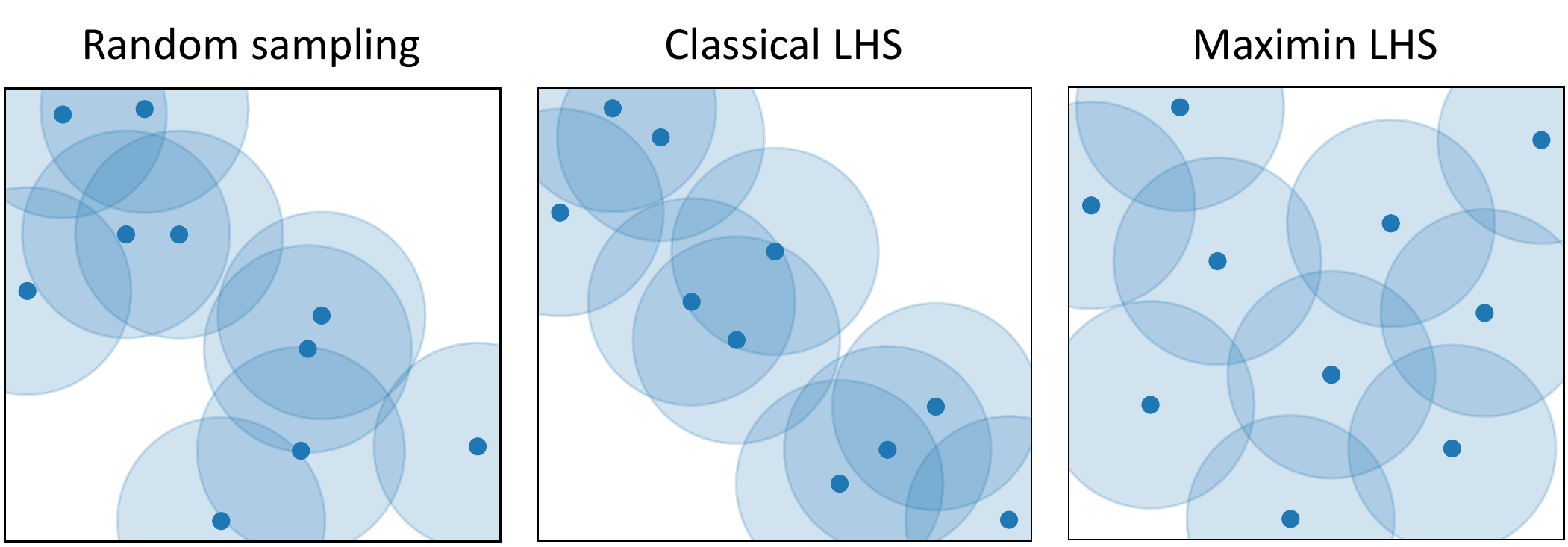}
    \caption{{Space-filling sampling}. 10 random samples are used to illustrate three different sampling strategies: (a) simple random sampling (SRS), (b) classical LHS, and (c) maximin LHS.}
    \label{fig:lhs}
        \vspace{-0.5cm}
\end{figure}

In flow-based models, the data probability density $p_{\mathcal{X}}(x)$ and latent density $p_{\mathcal{Z}}(z)$ follow the principle of probability preservation based on the change of variable theorem, which means the probability and statistical information are preserved in the transformation process \citep{li2006probability}. To this end, the statistics of prior $p(\mathbf{z})$ are preservably propagated to the posterior density $p(\mathbf{x}|\mathbf{y})$. Inspired by this observation, we propose to better manipulate the prior samples by introducing a space-filling sampling for latent space $\mathbf z$ such that a diverse set of solutions are fully explored. 

Instead of simple random sampling (SRS), we propose to use Latin Hypercube Sampling (LHS) \citep{stein1987large, shields2016generalization}, which is a variance-reduced sampling method and often used for Monte Carlo integration \citep{mckay2000comparison} and simulation \cite{zhang2021modern}. As shown in Fig.~\ref{fig:lhs}, the LHS design shows better performance on space-filling than SRS, specifically the optimized LHS with maximin criteria, where an LHS design $\mathbf Z_n = \left\{\mathbf z_1, ...,\mathbf z_n \right\}$ that maximizes the minimum distance between all pairs of points,  
\begin{equation}
    \mathbf Z_n = \argmax_{\mathbf Z_n} \min\left\{d(\mathbf z_i, \mathbf z_j): i \neq j =1,...,m \right\} \label{eq:lhs}
\end{equation}
where $d$ is the Euclidean distance defined by $d(\mathbf z, \mathbf z^{\prime}) = \sum_{j=1}^{m} (\mathbf z_j-\mathbf z_j^{\prime})^2$. Unlike quasi-Monte Carlo (QMC) methods \citep{caflisch1998monte}, e.g., Sobol sequence, that are limited in high dimensional problem \citep{kucherenko2015exploring}, maximin LHS works well with strong space-filling property and variance reduction capability. 

\vspace{0.2cm}
\noindent {\bf \texttt{iPage}: Accelerating Inverse Learning Process}. We propose an efficient learning algorithm for solving inverse learning problems by leveraging \underline{i}ntelligent \underline{p}rior with \underline{a}ccelerated \underline{g}radient-based \underline{e}stimate, with exploratory latent space sampling, which consists of three core steps: training, inference, and localization process, as explained in Algorithm 1.  

\begin{algorithm}[h!]\label{algo:1}
\footnotesize
  \caption{\hspace{-0.1cm}: iPage algorithm}
\begin{algorithmic}[1]
\STATE{\bf Require}: training data $\left\{(\mathbf{x}_i, \mathbf{y}_i)_{i=1}^m \right\}$, invertible neural network model $f(\mathbf{x};\phi)$, prior distribution $p(\mathbf{x})$,  
\vspace{0.1cm}
\STATE{// \bf \texttt{Training Process}}:
\vspace{0.1cm}
\STATE Initialize weight coefficients $\lambda_x$, $\lambda_y$ and $\lambda_z$  for each loss defined in Eq.~\eqref{eq:loss_y}-\eqref{eq:loss_x}  
\STATE Define an adaptive decay scheme for $\lambda_x$, $\lambda_y$ and $\lambda_z$ during the dynamic bi-directional training 
\STATE Minimize the total loss in Eq.~\eqref{eq:total_loss} via dynamic bi-directional training with gradient descent optimizer
\STATE Save the forward model $\hat{f}(\mathbf{x}; \phi^*)$ with the minimal $\ell_2$ loss 
\vspace{0.1cm}
\STATE{// \bf \texttt{Inference Process}}:
\vspace{0.1cm}
\STATE Generate a random sample $\mathbf{z}$ from latent space $p(\mathbf z)$ using optimized LHS with maximin criteria 
\STATE Compute the corresponding posterior sample $\hat{\mathbf{x}} = f^{-1}(\hat{\mathbf{y}}, \mathbf{z}; \phi)$ conditioned on the prior sample $\mathbf z$ and a specific observation $\hat{\mathbf{y}}$ through an invertible transformation 
\STATE Repeat sampling $\left\{\mathbf{z}_i \sim p(\mathbf{z}) \right\}_{i=1}^m$ to produce a number of posterior samples $\left\{\hat{\mathbf{x}}_i \sim \hat{p}({\mathbf{x}} | \hat{\mathbf{y}}) \right\}_{i=1}^m$ that follow the approximated posterior distribution $\hat{p}(\mathbf{x} | \hat{\mathbf{y}})$
\vspace{0.1cm}
\STATE{// \bf \texttt{Localization Process}}: 
\vspace{0.1cm}
\STATE Identify posterior samples $\left\{\hat{\mathbf{x}}_i \sim \hat{p}({\mathbf{x}} | \hat{\mathbf{y}}) \right\}_{i=1}^m$ as intelligent prior initialization to narrowing down the search space 
\STATE Compute the gradient $\mathbf{g}_i^{k}$ at the current $\hat{\mathbf{x}}_i$ using $\hat{f}_{\mathbf{y}}(\hat{\mathbf{x}}_i;{\phi}^*)$ in Eq.~\eqref{eq:gradient} using automatic differentiation. 
\STATE Localize these posterior samples precisely to exact solutions via gradient descent $\hat{\mathbf{x}}_{i}^{k+1} = \hat{\mathbf{x}}_i^{k} -\gamma ~ \mathbf{g}_i^{k}$
\STATE Return all possible exact inverse solutions $\mathbf{x}_i^{*}$
\end{algorithmic}
\end{algorithm}

\section{Experiments}
We start by demonstrating our proposed iPage method on a 2D sinewave function task and then extend our experiments to two artificial benchmark tasks. Finally, we introduce two real-world design problems from the field of quantum chemistry and additive manufacturing to illustrate the performance of learning complex high-dimensional inverse problems with practical objectives in natural sciences and engineering. 

\vspace{0.2cm}
\noindent{\bf Baseline Methods.} We provide five baseline methods: (1) Mixture density networks (MDN) \citep{bishop1994mixture}, which models the posterior distribution $p(\mathbf{x}|\mathbf{y})$ using a mixture of Gaussian models; (2) Invertible neural network (INN) \citep{ardizzone2018analyzing}, which is built on the flow-based model with latent variables to infer the completely posterior distribution; (3) Conditional invertible neural network (cINN) \citep{ardizzone2019guided, rombach2020network} which modifies INN framework by mapping the parameter space $\mathbf{x}$ and latent space $\mathbf{z}$ conditional on $\mathbf{y}$; (4) Conditional variational auto-encoder (cVAE) \citep{sohn2015learning}, which encodes $\mathbf{x}$ conditional on $\mathbf{y}$, into latent variables $\mathbf{z}$ based on the VAE framework, and (5) Neural-adjoint (NA) \cite{ren2020benchmarking}, which directly searches the global space
via surrogate-based optimization. To perform a fair comparison of all methods, we adjust neural network architectures such that all models have roughly the same number of model parameters. More information on benchmark details, baseline methods, invertible architectures, and datasets can be found in the Appendix.

\vspace{0.2cm}
\noindent{\bf Quantitative Metric.} We evaluate the true forward model $\Omega(\mathbf{x})$ at the generated inverse solutions $\mathbf{x}$ and measure the re-simulation error, which is defined as the mean squared error (MSE) to the target $\mathbf{y}^*$, $\mathcal{Q}_{\textup{re-sim}} = \mathbb{E}_{\mathbf{x}}\left\{ ||\Omega(\mathbf{x}) - \mathbf{y}^*||_2^2 \right\}$.
We apply this metric on two different scenarios: (1) solutions given 1000 different observations $\mathbf{y}_i^*, i=1,...,1000$, as shown in Table \ref{tab:1000_y}); and (2) solutions given a single specific observation $\mathbf{y}^*$, as shown in Table \ref{tab:1_y}.
\begin{table}[h!]
\vspace{-0.2cm}
\footnotesize
\centering
\caption{Training dataset size, input/output dimensionality, and target observation $y^*$ for benchmark tasks and real-world applications.} 
\vspace{-0.2cm}
\label{tab:dataset}
\begin{tabular}{@{}ccccc@{}}
\toprule
Task        & Dim $\mathbf{x}$ & Dim $\mathbf{y}$ & Data size & Target  $y^*$ \\ \midrule
Sinewave        & 2     & 1         & 1.00E+04  & $y^*=1.2$     \\
Robotic Arm & 4     & 2         & 1.00E+04 & $y^*=[1.5,0]$     \\
Ballistics  & 4     & 1          & 1.00E+04 & $y^*=5$     \\
Crystal     & 6     & 1          & 5.00E+03  & $y^*=0.5$    \\ 
Architecture     & 1024     & 1          & 1.00E+05  & $y^*=1.0$    \\ 
\bottomrule
\end{tabular}
\end{table}

\vspace{0.2cm}
\noindent{\bf Datasets.} We focus on five datasets, including three benchmarks (sinewave, robotic arm, and ballistics) and two real-world (crystal and architecture design) problems. As shown in Table \ref{tab:dataset}, the first four tasks are low-dimensional problems and the last one, i.e., the architecture task is a high-dimensional design problem in image pixel levels. 

\begin{table*}[!htbp]
\footnotesize
\centering
\caption{Performance comparison of tested methods on five tasks given 1000 different observations $\mathbf{y}^*$). The re-simulation error measures how well the generated $\hat{\mathbf{x}}$ is conditioned on the observation $\mathbf{y}^*$. Each task is performed 50 times to obtain the standard deviation.}
\vspace{-0.2cm}
\label{tab:1000_y}
\begin{tabular}{@{}cccccc@{}}
\toprule
Method & Sinewave     & Robotic Arm     & Ballistics     & Crystal Design   & Architecture Design                          \\ \midrule
Mixture density networks (MDN)    & 0.17 $\pm$ 2.3e-4  & 0.018  $\pm$  1.1e-5  & 0.024 $\pm$ 1.3e-5 & 0.81 $\pm$   2.3e-2  & 1.74 $\pm$ 2.5e-1   \\
Invertible neural network (INN)    & 0.12 $\pm$ 7.8e-5  & 0.014  $\pm$  8.2e-6  & 0.019 $\pm$ 9.9e-6 & 0.49 $\pm$ 3.9e-2   & 0.88 $\pm$ 9.7e-2 \\
conditional INN (cINN)   & 0.11 $\pm$ 2.3e-4  & 0.009 $\pm$ 7.3e-6    & 0.421 $\pm$ 2.0e-5   & 0.35 $\pm$ 8.1e-2  & 0.76 $\pm$ 8.6e-2    \\
conditional VAE (cVAE)   & 0.13 $\pm$ 3.9e-4  & 0.021 $\pm$  9.0e-6   & 0.798 $\pm$ 1.8e-5   & 0.64 $\pm$ 5.6e-2  & 1.03 $\pm$ 1.8e-1   \\
Neural-Adjoint (NA)     & 0.006 $\pm$ 4.1e-6 & 0.008  $\pm$  8.8e-6  & 0.016 $\pm$ 1.4e-5   & 0.12 $\pm$  {\bf 4.4e-3}  & 0.71 $\pm$ 1.1e-1 \\ \midrule
iPage (with maximin LHS)  & {\bf 0.002} $\pm$ {\bf 1.6e-6} & {\bf 0.006}  $\pm$  {\bf 4.2e-6}  & {\bf 0.011} $\pm$ {\bf 4.7e-6}   & {\bf 0.11} $\pm$ 4.5e-3  & {\bf 0.23} $\pm$ {\bf 1.2e-2} \\ \bottomrule
\end{tabular}
\end{table*}
\subsection{Illustrative Example: 2D Sinewave Function}
To test the capability of the iPage approach for solving inverse problems, we use a simple 2D sinusoidal function as a benchmark. The input parameters are $\mathbf{x} = [x_1, x_2]$ and the output is $y = \sin(3\pi x_1) + \cos(3\pi x_2)$. Due to its periodic nature, multiple solutions exist (theoretically infinite) given a specific observed $y^*$. 

\begin{figure}[h!]
    \centering
    \vspace{-0.2cm}
    \includegraphics[width=0.48\textwidth]{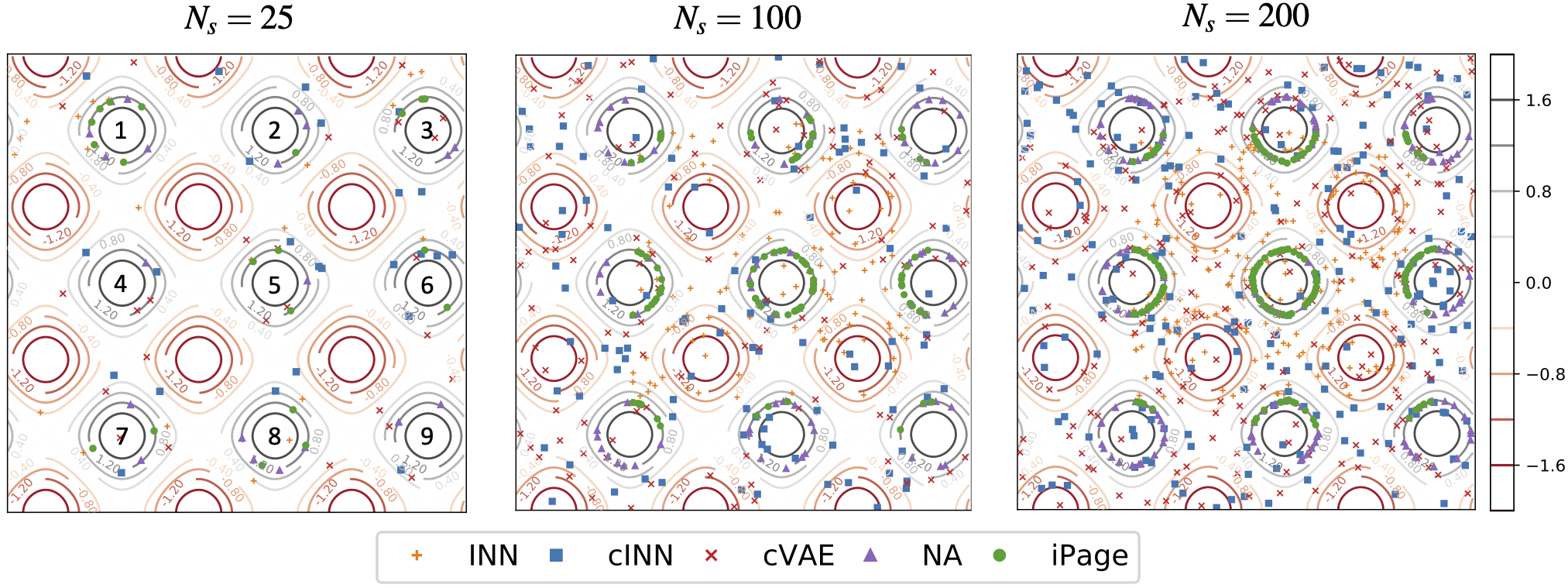}
    \caption{{Localization and exploration of inverse solutions for the 2D sinewave function}. Given a specific target $y^*=1.2$, there exits a multimodal disconnected solution space (labeled as 1-9 in the left panel). The inverse solution using four baseline methods (INN, cINN, cVAE, and NA) and iPage (with SRS) are illustrated and compared at different sampling counts ranging from 25 to 200.}
    \label{fig:sine9}
    \vspace{-0.2cm}
\end{figure}

For most of the existing baselines, this sinewave benchmark task remains a significant challenge task, specifically for obtaining accurate and diverse inverse solutions. An example of a fixed ${y}^*$ is shown in Fig.~\ref{fig:sine9}, where we compare our proposed methods to other baselines. We note that while the INN, cINN and cVAE methods are able to find some solutions within the local mode (marked by black circles labeled as 1-9), they fail to infer precise solutions. The NA method performs better in localizing to the globally optimal solution but fails to fully explore all possible solutions (e.g. missing mode 6). Our iPage method with simple random sampling (SRS) has the same difficulty (fails to capture modes 4 and 9) because the prior initialization fails to fully explore these local regions. Although this space exploration issue is mitigated by increasing the number of solutions as $N_s=100$, most of the localized solutions become concentrated on specific modes (e.g. modes 2, 5, and 6), with only limited solutions lie on the boundary modes (e.g. modes 9 and 3) for the case of $N_s=200$.

\begin{figure}[h!]
\vspace{-0.2cm}
    \centering
    \includegraphics[width=0.48\textwidth]{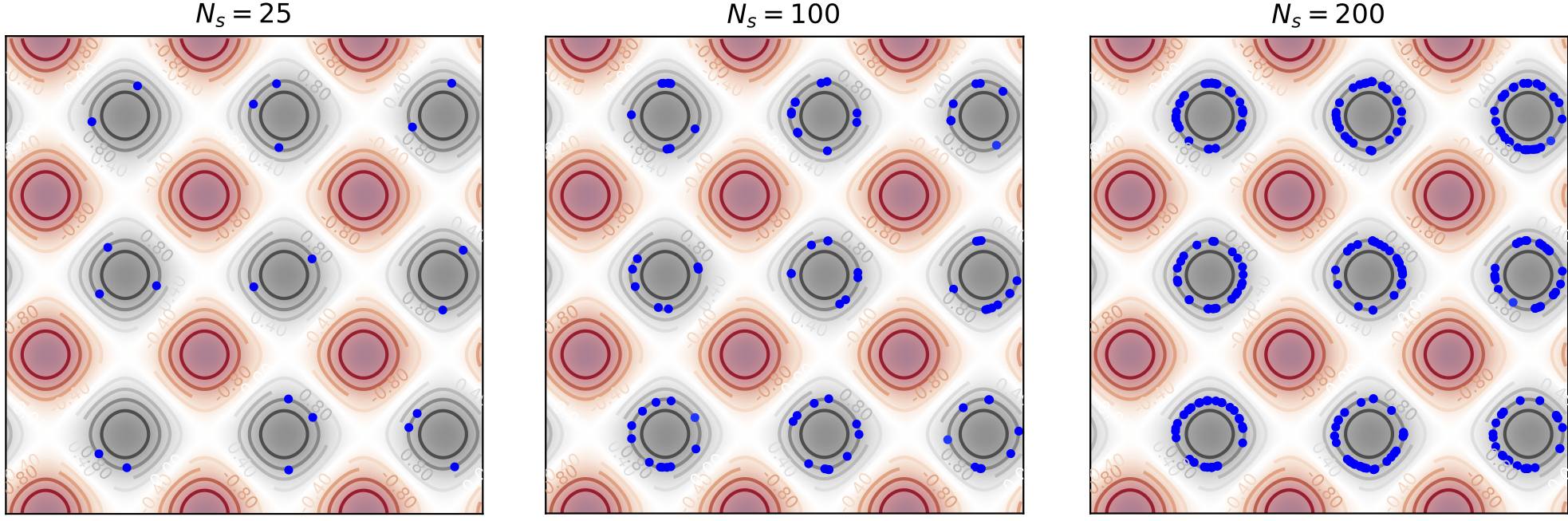}
    \caption{iPage (with mLHS) performance. Blue dots represent the final solutions, showing that our approach yields uniformly distributed solutions that capture all local modes.}
    \label{fig:iPage}
\end{figure}

To better capture all potential solutions, we introduce the iPage method with maximin LHS which leverages space-filling sampling to achieve better results than the previous models (see Fig.~\ref{fig:iPage}). All 9 local modes are evenly covered by the optimal solutions even with a limited number of samples (e.g., $N_s=25$). The quantitative comparison for the two scenarios is shown in Table \ref{tab:1000_y} and \ref{tab:1_y} respectively. iPage (with mLHS) shows superior performance, especially for the re-simulation error variance. This provides a clear illustration of the advantages of using a space-filling sampling for space exploration and variance reduction. 
\begin{figure*}[!h]
    \centering
    \includegraphics[width=0.9\textwidth]{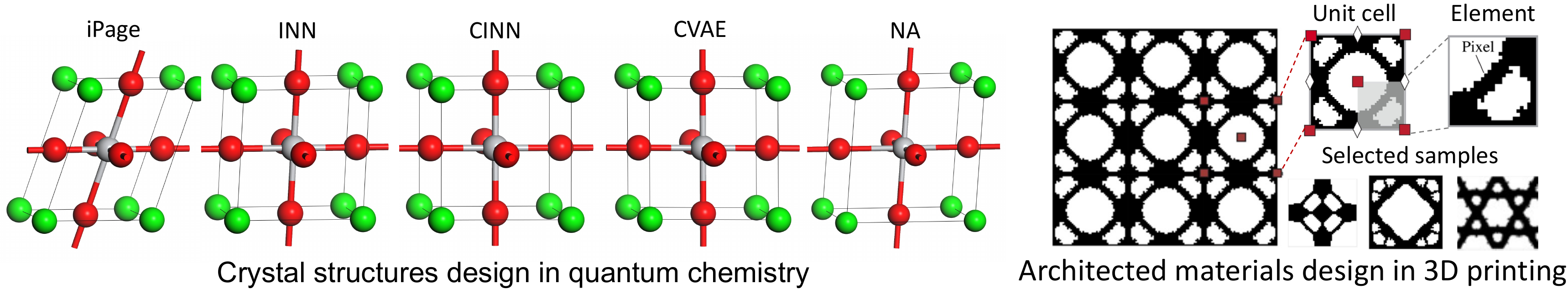}
    \caption{{Two real-world design applications}: (Left) Crystal structure design problem in quantum chemistry and (Right) Architected materials design problem in additive manufacturing.}
    \label{fig:crystal}
    \vspace{-0.2cm}
\end{figure*}

\subsection{Artificial Benchmark Tasks}
Two artificial benchmark tasks used by \citet{ardizzone2018analyzing,ren2020benchmarking, kruse2021benchmarking} are further used to assess the iPage performance.

\vspace{0.2cm}
\noindent{\bf Robotic Arm Task.} This is a geometric benchmark that targets the inference of the position of a multi-jointed robotic arm from various configurations of its joints. The inverse problem is to obtain all possible solutions in the $\mathbf{x}$-space given any observed 2D positions $\mathbf{y}^*$. For the case of multiple different observations, iPage shows similar results to cINN and NA but with a slightly lower variance, as shown in 
Table \ref{tab:1000_y}. In the second setting (see Table \ref{tab:1_y}), iPage outperforms the other baselines with a much lower error and variance.

\vspace{0.2cm}
\noindent{\bf Ballistics Task.} In this case, cINN and cVAE fail to solve the problem with much larger errors than the others while NA and INN show similar performance to iPage (see Table \ref{tab:1_y}). In general, iPage outperforms the baselines in terms of overall stability and robustness.    

\begin{figure*}[h!]
    \centering
    \vspace{-0.1cm}
    \includegraphics[width=0.9\textwidth]{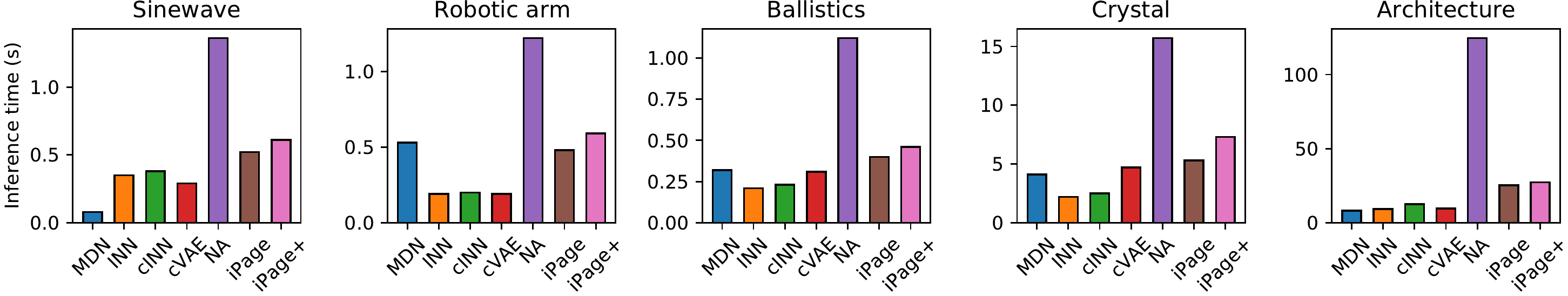}
    \vspace{-0.3cm}
    \caption{{Total time cost (inference and localization) for 1000 solutions}. The time-to-solution using iPage with other baselines on three benchmarks are compared side-by-side.}
    \label{fig:time}
        \vspace{-0.3cm}
\end{figure*}

\subsection{Real-world Applications}
We further demonstrate iPage's superiority in both natural sciences and engineering applications. 

\vspace{0.2cm}
\noindent{\bf Crystal Design Problem.} We apply our approach to a challenging real-world application in materials design, specifically one for modeling the electronic properties of complex metal oxides. 
Quantum chemistry simulations are performed to simulate these materials under perturbation and obtain their resulting electronic properties such as the band gap. Here, we tackle the inverse problem of the band gap to strain mapping for the case of the SrTiO\textsubscript{3} perovskite oxide, which is otherwise intractable to obtain from quantum chemistry. This can be an exceptionally difficult problem due to the complex underlying physics, and the high degree of sensitivity of the band gap to the lattice parameters, requiring very accurate predictions for the generated structures to succeed. 

\begin{table*}[!h] 
\footnotesize
\centering
\caption{Performance comparison of tested methods on five tasks for 1000 solutions conditioned on a specific observation $\mathbf{y}^*$. We repeat 50 times to obtain the standard deviation for each case.} 
\vspace{-0.2cm}
\label{tab:1_y}
\begin{tabular}{@{}cccccc@{}}
\toprule
Method & Sinewave     & Robotic Arm     & Ballistics     & Crystal Design  & Architecture Design \\ \midrule
Mixture density networks (MDN)    & 0.22 $\pm$ 5.1e-4 & 0.023 $\pm$ 2.3e-5  & 0.041 $\pm$ 2.9e-5 & 0.84 $\pm$ 3.3e-2   & 1.81 $\pm$ 2.0e-1\\
Invertible neural network (INN)    & 0.19 $\pm$ 9.3e-5  & 0.015 $\pm$ 4.7e-5  & {0.024} $\pm$ 1.9e-5 & 0.57 $\pm$ 4.7e-2  & 0.83 $\pm$ 9.1e-2 \\
conditional INN (cINN)   & 0.16 $\pm$ 5.0e-4  & 0.032 $\pm$ 3.1e-5   & 0.652 $\pm$ 4.3e-5   & 0.42 $\pm$ 8.8e-2 & 0.82 $\pm$ 8.5e-2   \\
conditional VAE (cVAE)   & 0.25 $\pm$ 7.0e-4  & 0.021 $\pm$ 5.6e-5   & 0.912 $\pm$ 3.2e-5   & 0.70 $\pm$ 9.0e-2  & 1.20 $\pm$ 1.7e-1 \\
Neural-Adjoint (NA)     & 0.011 $\pm$ 9.1e-6 & 0.012 $\pm$ 4.8e-5  & 0.031 $\pm$ 4.7e-5   & {0.15} $\pm$ 6.6e-3  & 0.79 $\pm$ 9.3e-2 \\ \midrule
iPage (with maximin LHS) & {\bf 0.004} $\pm$ {\bf 2.1e-6} & {\bf0.008} $\pm$ {\bf7.6e-6}  & {\bf 0.023} $\pm$ {\bf 8.9e-6}   & {\bf0.14} $\pm$ {\bf 2.2e-3}  & {\bf 0.22} $\pm$ {\bf 1.2e-2}  \\ \bottomrule
\end{tabular}
\end{table*}

Strain is represented by changes in the crystal lattice constants and angles, $a, b, c, \alpha, \beta, \gamma$, which serve as the hidden parameters $\mathbf{x}$. The target property y is the electronic band gap. 5000 samples were used for training where band gaps were obtained using quantum chemistry, representing the forward process. Additional details can be found in the appendix. We provide this dataset as a novel benchmark for computational chemistry, the first such example for solid state materials. We herein select an arbitrary target of 0.5 eV to generate our structures and compare the performance of our model with the existing ones. The new crystals are generated for each model and the band gaps are then computed using quantum chemistry for validation (see Fig.\ref{fig:crystal}). The performance of our approach was found to be significantly better than the baseline invertible models, INN, cINN, and cVAE, as shown in Table \ref{tab:1000_y} and \ref{tab:1_y}. By comparison, the INN, cINN, and cVAE models are consistently off the target by a far greater degree, with a deviation of 0.5-1.0 eV. The generated lattice parameters do not deviate much from the equilibrium values (see Fig.~\ref{fig:crystal}) and thus the results are unsurprisingly poor. Based on these observations and the magnitude of the deviations, it is unlikely these methods will provide useful results even with further training data provided. Only the NA method provides results with similar performance to iPage, though at a significantly greater computational cost. Furthermore, the NA method encounters difficulties for problems with a larger dimensionality in the parameter space. 

\vspace{0.2cm}
\noindent{\bf Architecture Materials Design Problem.} Architected materials on length scales from nanometers to meters are desirable for diverse applications \citep{mao2020designing}. Recent advances in additive manufacturing have made mass production of complex architected materials technologically and economically feasible. This task aims to find the optimal material layout by searching the design space (1024 dimensions) given a specific target mechanical property (see more details in the Appendix). The input is the pixel matrix for the element, and the output is the effective Young's modulus. We use this example to demonstrate that iPage can well scale to high-dimensional problems, e.g., pixel-level images, and outperform the other baseline methods. 

\subsection{Computational Cost Comparison}
We have demonstrated that the iPage can precisely localize the exact inverse solutions and quantitatively outperform INN, cINN, cVAE and MDN methods on five tasks. The NA method has advantages in learning accuracy but shows an obvious drawback of large computational costs compared to the other models. Fig.~\ref{fig:time} shows the total time cost including the inference and localization process on five tasks using one NVIDIA V100 GPU. Due to the invertible architecture, INN and cINN are efficient at sampling the posterior distributions. The time cost of iPage is slightly higher than INN, cINN and cVAE but still significantly lower than NA even though gradient descent is employed (few steps in local search). 

\section{Conclusion}
In this work, we develop an efficient inverse learning approach that utilizes posterior samples to accelerate the localization of all inverse solutions via gradient descent. To fully explore the parameter space, variance-reduced sampling strategies are imposed on the latent space to improve space-filling capability. Multiple experiments demonstrate that our approach outperforms the baselines and significantly improves the accuracy, efficiency, and robustness for solving inverse  problems, specifically in complex natural science and engineering design applications.  One current limitation is the efficiency of space-filling sampling in high-dimensional spaces. 
Future work will aim to improve sampling efficiency by leveraging scalable numerical algorithms. Also, we plan to apply the iPage method to broader topics in safe and robust AI, e.g., safe decision-making with Bayesian optimal experimental design \cite{zhang2021scalable}, and privacy defense in federated learning \cite{li2022auditing}.  

\bibliography{aaai23}

\begin{thebibliography}{66}
\providecommand{\natexlab}[1]{#1}

\bibitem[{Ardizzone et~al.(2018)Ardizzone, Kruse, Wirkert, Rahner, Pellegrini,
  Klessen, Maier-Hein, Rother, and K{\"o}the}]{ardizzone2018analyzing}
Ardizzone, L.; Kruse, J.; Wirkert, S.; Rahner, D.; Pellegrini, E.~W.; Klessen,
  R.~S.; Maier-Hein, L.; Rother, C.; and K{\"o}the, U. 2018.
\newblock Analyzing inverse problems with invertible neural networks.
\newblock \emph{arXiv preprint arXiv:1808.04730}.

\bibitem[{Ardizzone et~al.(2019)Ardizzone, L{\"u}th, Kruse, Rother, and
  K{\"o}the}]{ardizzone2019guided}
Ardizzone, L.; L{\"u}th, C.; Kruse, J.; Rother, C.; and K{\"o}the, U. 2019.
\newblock Guided image generation with conditional invertible neural networks.
\newblock \emph{arXiv preprint arXiv:1907.02392}.

\bibitem[{Asim et~al.(2020)Asim, Daniels, Leong, Ahmed, and
  Hand}]{asim2020invertible}
Asim, M.; Daniels, M.; Leong, O.; Ahmed, A.; and Hand, P. 2020.
\newblock Invertible generative models for inverse problems: mitigating
  representation error and dataset bias.
\newblock In \emph{International Conference on Machine Learning}, 399--409.
  PMLR.

\bibitem[{Bendsoe and Sigmund(2013)}]{bendsoe2013topology}
Bendsoe, M.~P.; and Sigmund, O. 2013.
\newblock \emph{Topology optimization: theory, methods, and applications}.
\newblock Springer Science \& Business Media.

\bibitem[{Bishop(1994)}]{bishop1994mixture}
Bishop, C.~M. 1994.
\newblock Mixture density networks.

\bibitem[{Bishop(2006)}]{bishop2006pattern}
Bishop, C.~M. 2006.
\newblock \emph{Pattern recognition and machine learning}.
\newblock springer.

\bibitem[{Bl{\"o}chl(1994)}]{RN141}
Bl{\"o}chl, P.~E. 1994.
\newblock Projector augmented-wave method.
\newblock \emph{Physical review B}, 50(24): 17953.

\bibitem[{Caflisch et~al.(1998)}]{caflisch1998monte}
Caflisch, R.~E.; et~al. 1998.
\newblock Monte carlo and quasi-monte carlo methods.
\newblock \emph{Acta numerica}, 1998: 1--49.

\bibitem[{Daras et~al.(2021)Daras, Dean, Jalal, and
  Dimakis}]{daras2021intermediate}
Daras, G.; Dean, J.; Jalal, A.; and Dimakis, A.~G. 2021.
\newblock Intermediate layer optimization for inverse problems using deep
  generative models.
\newblock \emph{arXiv preprint arXiv:2102.07364}.

\bibitem[{Deng et~al.(2021)Deng, Ren, Fan, Malof, and Padilla}]{deng2021neural}
Deng, Y.; Ren, S.; Fan, K.; Malof, J.~M.; and Padilla, W.~J. 2021.
\newblock Neural-adjoint method for the inverse design of all-dielectric
  metasurfaces.
\newblock \emph{Optics Express}, 29(5): 7526--7534.

\bibitem[{Dinh, Sohl-Dickstein, and Bengio(2016)}]{dinh2016density}
Dinh, L.; Sohl-Dickstein, J.; and Bengio, S. 2016.
\newblock Density estimation using real nvp.
\newblock \emph{arXiv preprint arXiv:1605.08803}.

\bibitem[{Dudarev et~al.(1998)Dudarev, Botton, Savrasov, Humphreys, and
  Sutton}]{RN147}
Dudarev, S.; Botton, G.; Savrasov, S.; Humphreys, C.; and Sutton, A. 1998.
\newblock Electron-energy-loss spectra and the structural stability of nickel
  oxide: An LSDA+ U study.
\newblock \emph{Physical Review B}, 57(3): 1505.

\bibitem[{Durkan et~al.(2019)Durkan, Bekasov, Murray, and
  Papamakarios}]{durkan2019neural}
Durkan, C.; Bekasov, A.; Murray, I.; and Papamakarios, G. 2019.
\newblock Neural spline flows.
\newblock \emph{arXiv preprint arXiv:1906.04032}.

\bibitem[{Ertl and Schuffenhauer(2009)}]{ertl2009estimation}
Ertl, P.; and Schuffenhauer, A. 2009.
\newblock Estimation of synthetic accessibility score of drug-like molecules
  based on molecular complexity and fragment contributions.
\newblock \emph{Journal of cheminformatics}, 1(1): 1--11.

\bibitem[{Forrester and Keane(2009)}]{forrester2009recent}
Forrester, A.~I.; and Keane, A.~J. 2009.
\newblock Recent advances in surrogate-based optimization.
\newblock \emph{Progress in aerospace sciences}, 45(1-3): 50--79.

\bibitem[{Fung et~al.(2022)Fung, Jia, Zhang, Bi, Yin, and
  Ganesh}]{fung2022atomic}
Fung, V.; Jia, S.; Zhang, J.; Bi, S.; Yin, J.; and Ganesh, P. 2022.
\newblock Atomic structure generation from reconstructing structural
  fingerprints.
\newblock \emph{Machine Learning: Science and Technology}.

\bibitem[{Fung et~al.(2021)Fung, Zhang, Hu, Ganesh, and
  Sumpter}]{fung2021inverse}
Fung, V.; Zhang, J.; Hu, G.; Ganesh, P.; and Sumpter, B.~G. 2021.
\newblock Inverse design of two-dimensional materials with invertible neural
  networks.
\newblock \emph{npj Computational Materials}, 7(1): 1--9.

\bibitem[{G{\'o}mez-Bombarelli et~al.(2018)G{\'o}mez-Bombarelli, Wei, Duvenaud,
  Hern{\'a}ndez-Lobato, S{\'a}nchez-Lengeling, Sheberla, Aguilera-Iparraguirre,
  Hirzel, Adams, and Aspuru-Guzik}]{gomez2018automatic}
G{\'o}mez-Bombarelli, R.; Wei, J.~N.; Duvenaud, D.; Hern{\'a}ndez-Lobato,
  J.~M.; S{\'a}nchez-Lengeling, B.; Sheberla, D.; Aguilera-Iparraguirre, J.;
  Hirzel, T.~D.; Adams, R.~P.; and Aspuru-Guzik, A. 2018.
\newblock Automatic chemical design using a data-driven continuous
  representation of molecules.
\newblock \emph{ACS central science}, 4(2): 268--276.

\bibitem[{Grathwohl et~al.(2018)Grathwohl, Chen, Bettencourt, Sutskever, and
  Duvenaud}]{grathwohl2018ffjord}
Grathwohl, W.; Chen, R.~T.; Bettencourt, J.; Sutskever, I.; and Duvenaud, D.
  2018.
\newblock Ffjord: Free-form continuous dynamics for scalable reversible
  generative models.
\newblock \emph{arXiv preprint arXiv:1810.01367}.

\bibitem[{Gretton et~al.(2012{\natexlab{a}})Gretton, Borgwardt, Rasch,
  Sch{\"o}lkopf, and Smola}]{gretton2012kernel}
Gretton, A.; Borgwardt, K.~M.; Rasch, M.~J.; Sch{\"o}lkopf, B.; and Smola, A.
  2012{\natexlab{a}}.
\newblock A kernel two-sample test.
\newblock \emph{The Journal of Machine Learning Research}, 13(1): 723--773.

\bibitem[{Gretton et~al.(2012{\natexlab{b}})Gretton, Sejdinovic, Strathmann,
  Balakrishnan, Pontil, Fukumizu, and Sriperumbudur}]{gretton2012optimal}
Gretton, A.; Sejdinovic, D.; Strathmann, H.; Balakrishnan, S.; Pontil, M.;
  Fukumizu, K.; and Sriperumbudur, B.~K. 2012{\natexlab{b}}.
\newblock Optimal kernel choice for large-scale two-sample tests.
\newblock In \emph{Advances in neural information processing systems},
  1205--1213. Citeseer.

\bibitem[{Kingma and Dhariwal(2018)}]{kingma2018glow}
Kingma, D.~P.; and Dhariwal, P. 2018.
\newblock Glow: Generative flow with invertible 1x1 convolutions.
\newblock \emph{arXiv preprint arXiv:1807.03039}.

\bibitem[{Kothari et~al.(2021)Kothari, Khorashadizadeh, Maarten, and
  Dokmanic}]{kothari2021trumpets}
Kothari, K.; Khorashadizadeh, A.; Maarten, V.; and Dokmanic, I. 2021.
\newblock Trumpets: Injective Flows for Inference and Inverse Problems.

\bibitem[{Kresse and Furthm{\"u}ller(1996{\natexlab{a}})}]{RN138}
Kresse, G.; and Furthm{\"u}ller, J. 1996{\natexlab{a}}.
\newblock Efficiency of ab-initio total energy calculations for metals and
  semiconductors using a plane-wave basis set.
\newblock \emph{Computational materials science}, 6(1): 15--50.

\bibitem[{Kresse and Furthm{\"u}ller(1996{\natexlab{b}})}]{RN144}
Kresse, G.; and Furthm{\"u}ller, J. 1996{\natexlab{b}}.
\newblock Efficient iterative schemes for ab initio total-energy calculations
  using a plane-wave basis set.
\newblock \emph{Physical review B}, 54(16): 11169.

\bibitem[{Kruse et~al.(2021)Kruse, Ardizzone, Rother, and
  K{\"o}the}]{kruse2021benchmarking}
Kruse, J.; Ardizzone, L.; Rother, C.; and K{\"o}the, U. 2021.
\newblock Benchmarking invertible architectures on inverse problems.
\newblock \emph{arXiv preprint arXiv:2101.10763}.

\bibitem[{Kucherenko, Albrecht, and Saltelli(2015)}]{kucherenko2015exploring}
Kucherenko, S.; Albrecht, D.; and Saltelli, A. 2015.
\newblock Exploring multi-dimensional spaces: A comparison of Latin hypercube
  and quasi Monte Carlo sampling techniques.
\newblock \emph{arXiv preprint arXiv:1505.02350}.

\bibitem[{Lavin et~al.(2021)Lavin, Zenil, Paige, Krakauer, Gottschlich,
  Mattson, Anandkumar, Choudry, Rocki, Baydin et~al.}]{lavin2021simulation}
Lavin, A.; Zenil, H.; Paige, B.; Krakauer, D.; Gottschlich, J.; Mattson, T.;
  Anandkumar, A.; Choudry, S.; Rocki, K.; Baydin, A.~G.; et~al. 2021.
\newblock Simulation intelligence: Towards a new generation of scientific
  methods.
\newblock \emph{arXiv preprint arXiv:2112.03235}.

\bibitem[{Li and Chen(2006)}]{li2006probability}
Li, J.; and Chen, J.-B. 2006.
\newblock The probability density evolution method for dynamic response
  analysis of non-linear stochastic structures.
\newblock \emph{International Journal for Numerical Methods in Engineering},
  65(6): 882--903.

\bibitem[{Li et~al.(2022)Li, Zhang, Liu, and Liu}]{li2022auditing}
Li, Z.; Zhang, J.; Liu, L.; and Liu, J. 2022.
\newblock Auditing Privacy Defenses in Federated Learning via Generative
  Gradient Leakage.
\newblock In \emph{Proceedings of the IEEE/CVF Conference on Computer Vision
  and Pattern Recognition}, 10132--10142.

\bibitem[{Liu, Zhang, and Pei(2022)}]{liu2022machine}
Liu, X.; Zhang, J.; and Pei, Z. 2022.
\newblock Machine learning for high-entropy alloys: Progress, challenges and
  opportunities.
\newblock \emph{Progress in Materials Science}, 101018.

\bibitem[{Ma et~al.(2019)Ma, Cheng, Xu, Wen, and Liu}]{ma2019probabilistic}
Ma, W.; Cheng, F.; Xu, Y.; Wen, Q.; and Liu, Y. 2019.
\newblock Probabilistic representation and inverse design of metamaterials
  based on a deep generative model with semi-supervised learning strategy.
\newblock \emph{Advanced Materials}, 31(35): 1901111.

\bibitem[{Mao, He, and Zhao(2020)}]{mao2020designing}
Mao, Y.; He, Q.; and Zhao, X. 2020.
\newblock Designing complex architectured materials with generative adversarial
  networks.
\newblock \emph{Science Advances}, 6(17): eaaz4169.

\bibitem[{McKay, Beckman, and Conover(2000)}]{mckay2000comparison}
McKay, M.~D.; Beckman, R.~J.; and Conover, W.~J. 2000.
\newblock A comparison of three methods for selecting values of input variables
  in the analysis of output from a computer code.
\newblock \emph{Technometrics}, 42(1): 55--61.

\bibitem[{Monkhorst and Pack(1976)}]{RN148}
Monkhorst, H.~J.; and Pack, J.~D. 1976.
\newblock Special points for Brillouin-zone integrations.
\newblock \emph{Physical review B}, 13(12): 5188.

\bibitem[{Nielsen et~al.(2020)Nielsen, Jaini, Hoogeboom, Winther, and
  Welling}]{nielsen2020survae}
Nielsen, D.; Jaini, P.; Hoogeboom, E.; Winther, O.; and Welling, M. 2020.
\newblock Survae flows: Surjections to bridge the gap between vaes and flows.
\newblock \emph{Advances in Neural Information Processing Systems}, 33.

\bibitem[{Perdew, Burke, and Ernzerhof(1996)}]{RN145}
Perdew, J.~P.; Burke, K.; and Ernzerhof, M. 1996.
\newblock Generalized gradient approximation made simple.
\newblock \emph{Physical review letters}, 77(18): 3865.

\bibitem[{Ren, Padilla, and Malof(2020)}]{ren2020benchmarking}
Ren, S.; Padilla, W.; and Malof, J. 2020.
\newblock Benchmarking deep inverse models over time, and the neural-adjoint
  method.
\newblock \emph{arXiv preprint arXiv:2009.12919}.

\bibitem[{Rezende and Mohamed(2015)}]{rezende2015variational}
Rezende, D.; and Mohamed, S. 2015.
\newblock Variational inference with normalizing flows.
\newblock In \emph{International Conference on Machine Learning}, 1530--1538.
  PMLR.

\bibitem[{Ricca et~al.(2020)Ricca, Timrov, Cococcioni, Marzari, and
  Aschauer}]{RN6481}
Ricca, C.; Timrov, I.; Cococcioni, M.; Marzari, N.; and Aschauer, U. 2020.
\newblock Self-consistent DFT+ U+ V study of oxygen vacancies in SrTiO 3.
\newblock \emph{Physical review research}, 2(2): 023313.

\bibitem[{Rombach, Esser, and Ommer(2020)}]{rombach2020network}
Rombach, R.; Esser, P.; and Ommer, B. 2020.
\newblock Network-to-Network Translation with Conditional Invertible Neural
  Networks.
\newblock \emph{Advances in Neural Information Processing Systems}, 33.

\bibitem[{Sanchez-Lengeling and Aspuru-Guzik(2018)}]{sanchez2018inverse}
Sanchez-Lengeling, B.; and Aspuru-Guzik, A. 2018.
\newblock Inverse molecular design using machine learning: Generative models
  for matter engineering.
\newblock \emph{Science}, 361(6400): 360--365.

\bibitem[{Shields and Zhang(2016)}]{shields2016generalization}
Shields, M.~D.; and Zhang, J. 2016.
\newblock The generalization of Latin hypercube sampling.
\newblock \emph{Reliability Engineering \& System Safety}, 148: 96--108.

\bibitem[{Sohn, Lee, and Yan(2015)}]{sohn2015learning}
Sohn, K.; Lee, H.; and Yan, X. 2015.
\newblock Learning structured output representation using deep conditional
  generative models.
\newblock \emph{Advances in neural information processing systems}, 28:
  3483--3491.

\bibitem[{Song et~al.(2021)Song, Shen, Xing, and Ermon}]{song2021solving}
Song, Y.; Shen, L.; Xing, L.; and Ermon, S. 2021.
\newblock Solving Inverse Problems in Medical Imaging with Score-Based
  Generative Models.
\newblock In \emph{International Conference on Learning Representations}.

\bibitem[{Stein(1987)}]{stein1987large}
Stein, M. 1987.
\newblock Large sample properties of simulations using Latin hypercube
  sampling.
\newblock \emph{Technometrics}, 29(2): 143--151.

\bibitem[{Sun and Bouman(2020)}]{sun2020deep}
Sun, H.; and Bouman, K.~L. 2020.
\newblock Deep Probabilistic Imaging: Uncertainty Quantification and
  Multi-modal Solution Characterization for Computational Imaging.
\newblock \emph{arXiv preprint arXiv:2010.14462}.

\bibitem[{Tokura, Kawasaki, and Nagaosa(2017)}]{RN6466}
Tokura, Y.; Kawasaki, M.; and Nagaosa, N. 2017.
\newblock Emergent functions of quantum materials.
\newblock \emph{Nature Physics}, 13(11): 1056--1068.

\bibitem[{Tonolini et~al.(2020)Tonolini, Radford, Turpin, Faccio, and
  Murray-Smith}]{tonolini2020variational}
Tonolini, F.; Radford, J.; Turpin, A.; Faccio, D.; and Murray-Smith, R. 2020.
\newblock Variational inference for computational imaging inverse problems.
\newblock \emph{Journal of Machine Learning Research}, 21(179): 1--46.

\bibitem[{Wang, Ye, and De~Man(2020)}]{wang2020deep}
Wang, G.; Ye, J.~C.; and De~Man, B. 2020.
\newblock Deep learning for tomographic image reconstruction.
\newblock \emph{Nature Machine Intelligence}, 2(12): 737--748.

\bibitem[{Wang et~al.(2018)Wang, Liu, Zhu, Tao, Kautz, and
  Catanzaro}]{wang2018high}
Wang, T.-C.; Liu, M.-Y.; Zhu, J.-Y.; Tao, A.; Kautz, J.; and Catanzaro, B.
  2018.
\newblock High-resolution image synthesis and semantic manipulation with
  conditional gans.
\newblock In \emph{Proceedings of the IEEE conference on computer vision and
  pattern recognition}, 8798--8807.

\bibitem[{Whang, Lei, and Dimakis(2021)}]{whang2021solving}
Whang, J.; Lei, Q.; and Dimakis, A. 2021.
\newblock Solving Inverse Problems with a Flow-based Noise Model.
\newblock In \emph{International Conference on Machine Learning}, 11146--11157.
  PMLR.

\bibitem[{Whang, Lindgren, and Dimakis(2021)}]{whang2021composing}
Whang, J.; Lindgren, E.; and Dimakis, A. 2021.
\newblock Composing Normalizing Flows for Inverse Problems.
\newblock In \emph{International Conference on Machine Learning}, 11158--11169.
  PMLR.

\bibitem[{White et~al.(2019)White, Arrighi, Kudo, and
  Watts}]{white2019multiscale}
White, D.~A.; Arrighi, W.~J.; Kudo, J.; and Watts, S.~E. 2019.
\newblock Multiscale topology optimization using neural network surrogate
  models.
\newblock \emph{Computer Methods in Applied Mechanics and Engineering}, 346:
  1118--1135.

\bibitem[{Wu, K{\"o}hler, and No{\'e}(2020)}]{wu2020stochastic}
Wu, H.; K{\"o}hler, J.; and No{\'e}, F. 2020.
\newblock Stochastic normalizing flows.
\newblock \emph{arXiv preprint arXiv:2002.06707}.

\bibitem[{Yang, Strukov, and Stewart(2013)}]{RN6431}
Yang, J.~J.; Strukov, D.~B.; and Stewart, D.~R. 2013.
\newblock Memristive devices for computing.
\newblock \emph{Nature Nanotechnology}, 8(1): 13--24.

\bibitem[{Yang et~al.(2018)Yang, Dai, Kiyavash, and He}]{yang2018predictive}
Yang, Y.; Dai, B.; Kiyavash, N.; and He, N. 2018.
\newblock Predictive approximate Bayesian computation via saddle points.
\newblock \emph{Proceedings of Machine Learning Research}.

\bibitem[{Yao et~al.(2021)Yao, S{\'a}nchez-Lengeling, Bobbitt, Bucior, Kumar,
  Collins, Burns, Woo, Farha, Snurr et~al.}]{yao2021inverse}
Yao, Z.; S{\'a}nchez-Lengeling, B.; Bobbitt, N.~S.; Bucior, B.~J.; Kumar, S.
  G.~H.; Collins, S.~P.; Burns, T.; Woo, T.~K.; Farha, O.~K.; Snurr, R.~Q.;
  et~al. 2021.
\newblock Inverse design of nanoporous crystalline reticular materials with
  deep generative models.
\newblock \emph{Nature Machine Intelligence}, 3(1): 76--86.

\bibitem[{Zhang, Zhang, and Hinkle(2019)}]{zhang2019learning}
Zhang, G.; Zhang, J.; and Hinkle, J. 2019.
\newblock Learning nonlinear level sets for dimensionality reduction in
  function approximation.
\newblock \emph{Advances in Neural Information Processing Systems}, 32.

\bibitem[{Zhang(2021)}]{zhang2021modern}
Zhang, J. 2021.
\newblock Modern Monte Carlo methods for efficient uncertainty quantification
  and propagation: A survey.
\newblock \emph{Wiley Interdisciplinary Reviews: Computational Statistics},
  13(5): e1539.

\bibitem[{Zhang, Bi, and Zhang(2021)}]{zhang2021scalable}
Zhang, J.; Bi, S.; and Zhang, G. 2021.
\newblock A Scalable Gradient Free Method for Bayesian Experimental Design with
  Implicit Models.
\newblock In \emph{International Conference on Artificial Intelligence and
  Statistics}, 3745--3753. PMLR.

\bibitem[{Zhang and Shields(2018{\natexlab{a}})}]{zhang2018effect}
Zhang, J.; and Shields, M.~D. 2018{\natexlab{a}}.
\newblock The effect of prior probabilities on quantification and propagation
  of imprecise probabilities resulting from small datasets.
\newblock \emph{Computer Methods in Applied Mechanics and Engineering}, 334:
  483--506.

\bibitem[{Zhang and Shields(2018{\natexlab{b}})}]{zhang2018quantification}
Zhang, J.; and Shields, M.~D. 2018{\natexlab{b}}.
\newblock On the quantification and efficient propagation of imprecise
  probabilities resulting from small datasets.
\newblock \emph{Mechanical Systems and Signal Processing}, 98: 465--483.

\bibitem[{Zhang et~al.(2021)Zhang, Tran, Lu, and Zhang}]{zhang2021enabling}
Zhang, J.; Tran, H.; Lu, D.; and Zhang, G. 2021.
\newblock Enabling long-range exploration in minimization of multimodal
  functions.
\newblock In \emph{Uncertainty in Artificial Intelligence}, 1639--1649. PMLR.

\bibitem[{Zhang et~al.(2016)Zhang, Liu, Zhuang, Kent, Cooper, Ganesh, and
  Xu}]{RN6480}
Zhang, L.; Liu, B.; Zhuang, H.; Kent, P.~R.; Cooper, V.~R.; Ganesh, P.; and Xu,
  H. 2016.
\newblock Oxygen vacancy diffusion in bulk SrTiO3 from density functional
  theory calculations.
\newblock \emph{Computational Materials Science}, 118: 309--315.

\bibitem[{Zhu et~al.(2020)Zhu, Zhang, Yang, and Huang}]{RN6467}
Zhu, J.; Zhang, T.; Yang, Y.; and Huang, R. 2020.
\newblock A comprehensive review on emerging artificial neuromorphic devices.
\newblock 7(1): 011312.

\end{thebibliography}

\appendix
\section{Appendix}
\section{Baseline Inverse Methods}
\label{Baseline Methods}
Here we provide more technical details about the baseline inverse methods used in our paper. 
\subsection{Invertible Neural Network (INN)}
This inverse model is our fundamental baseline, which is based on the invertible architecture with affine coupling layers. The basic idea is to define a forward L2 loss for fitting the y-prediction to the training data 
\begin{equation}
    \mathcal{L}_{\mathbf{y}} = ||\mathbf{y}-\mathbf{y}_t||_2^2
\end{equation}
where $\mathbf{y}_t$ is the true output. Then a backward MMD loss $\mathcal{L}_{\mathbf{z}}=\textup{MMD}(\mathbf{z})$ is used to fit the probability distribution of latent variable $p(\mathbf{z})$ to a standard Gaussian distribution $\mathcal{N}(\mathbf{0}, \mathbf{I})$. Therefore, a total loss is defined with weighting factors $\lambda_{\mathbf{y}}$ and $\lambda_{\mathbf{z}}$:
\begin{equation}
    \mathcal{L} = \lambda_{\mathbf{y}} \mathcal{L}_{\mathbf{y}} + \lambda_{\mathbf{z}} \mathcal{L}_{\mathbf{z}}
\end{equation}

Kruse et al.\cite{kruse2021benchmarking} proposed an alternative way to train the INNs with a maximum likelihood loss. This is achieved by assuming $\mathbf{y}$ to be normally distributed around the true values $\mathbf{y}_t$ with very low variance $\sigma^2$: 
\begin{equation}
    \mathcal{L} = \frac{1}{2} \left(\frac{1}{\sigma^2} \cdot (\mathbf{y} - \mathbf{y}_t)^2 + \mathbf{z}^2 \right) - \log |\textup{det} J_{\mathbf{x} \mapsto [\mathbf{y},\mathbf{z}]} |
\end{equation}

\subsection{Conditional Neural Network (cINN)}
The INN model may have a challenge when the dimensionality of $\mathbf{x}$ is significantly larger than $\mathbf{y}$, for example, in image-based inverse problems. Thus, instead of training INN to predict $\mathbf{y}$ and $\mathbf{x}$ with additional latent variable $\mathbf{z}$, cINN transforms $\mathbf{x}$ directly to a latent representation $\mathbf{z}$ conditional on the observation $\mathbf{y}$. This is achieved by using $\mathbf{y}$ as an additional input to each affine coupling layer in both the forward and backward processes. We can also use maximum likelihood to train cINN as 
\begin{equation}
    \mathcal{L} = \frac{1}{2} \cdot \mathbf{z}^2 - \log \left|\textup{det} J_{\mathbf{x} \mapsto \mathbf{z}} \right|
\end{equation}
We use the original authors' implementation in both invertible architecture \cite{kruse2021benchmarking}. 

\subsection{Conditional Variational Autoencoder (cVAE)}
The conditional variational autoencoder uses the evidence lower bound and encodes the $\mathbf{x}$ into Gaussian distributed random latent variable $\mathbf{z}$ conditioned on $\mathbf{y}$. The forward training process utilizes the L2 loss to achieve a good reconstruction of the original input $\mathbf{x}$, and the backward process is to solve $\mathbf{x}$ which is decoded from random samples that are drawn from latent space $\mathbf{z}$ conditioned on $\mathbf{y}$. The loss function is defined as 
\begin{equation}
    \mathcal{L} = \alpha \cdot (\mathbf{x}-\hat{\mathbf{x}})^2 - \frac{1}{2} \cdot \beta \cdot (1+\log \mathbf{\sigma}_{z} - \mathbf{\mu}_z^2 -\mathbf{\sigma}_z)
\end{equation}
We use the code implemented by \cite{ma2019probabilistic} for this method. 

\subsection{Mixture Density Network (MDN)}
The mixture density network can model the inverse problem but it is not an invertible model. MDN uses $\mathbf{y}$ as an input and predicts the parameters $\mathbf{\mu}_x, \Sigma_x^{-1}$ of a Gaussian mixture model $p(\mathbf{x}|\mathbf{y})$. We train the MDN model by maximizing the likelihood of the training data with the following loss function:
\begin{equation}
    \mathcal{L} = \frac{1}{2} \cdot \left( \mathbf{x} \mathbf{\mu}_x^{\top} \cdot \Sigma_x^{-1} \cdot \mathbf{x} \mathbf{\mu}_x \right) - \log |\Sigma_x^{-1}|^{1/2}
\end{equation}
We implemented this method according to \cite{bishop2006pattern}.

\subsection{The Neural-Adjoint (NA) Method}
The NA method, different from the other methods, uses a deep neural network as a surrogate (i.e., approximation) for the forward model and then uses backpropagation with respect to the input variable to search for good inverse solutions \cite{ren2020benchmarking}. The authors \cite{ren2020benchmarking} propose an alternative metric that quantifies the expected minimum error by drawing a sequence of $z$ values of length T, denoted $Z_T$, which is given as:
\begin{equation}
    r_T = \mathbb{E}_{(x,y) \sim \mathcal{D}, Z_T \sim \Omega} \left[ \min_{z \in Z_T} [ \mathcal{L}(\hat{y}(z)),y]\right]
\end{equation}
where $Z_T$ is a sequence of length $T$ drawn from a distribution $\Omega$, which characterizes the expected loss of an inverse problem as a function of the number of samples of $z$ for each target $y$. They claim that solving inverse problems strongly depends on $T$. Additionally, to narrow the stochastic search space in the model input domain, a boundary loss is proposed to ensure the NA identifies solutions that are within the training data domain, which is given by:
\begin{equation}
    \mathcal{L}_b = \texttt{ReLU}(|\hat{x} - \mu_x| - \frac{1}{2}R_x)
\end{equation}
where $R_x$ is the bounded range of input domain, $\mu_x$ is the mean of the training data, and $\texttt{ReLU}$ is the conventional nonlinear activation function. For this method, we use the implementation provided by the authors \cite{ren2020benchmarking} at \url{https://github.com/BensonRen/BDIMNNA}.

\section{Artificial Benchmark Tasks: Additional Details}
Two benchmark tasks that have been used by \citep{ardizzone2018analyzing,ren2020benchmarking, kruse2021benchmarking} are also employed to assess the algorithm performance. 

\subsection{Robotic arm task}
This is a geometric benchmark example that targets the inference of the position of a multi-jointed robotic arm from various configurations of its joints. There are four input parameters: starting height $x_1$, three joint angles $x_2$, $x_3$, and $x_4$ in the forward model. The output is the arm's position $[y_1, y_2]$ as: 
\begin{equation}
\begin{aligned}
    y_1 &= l_1\sin(x_2) + l_2\sin(x_3-x_2) \\
    &+l_3\sin(x_4-x_2-x_3) + x_1  
\end{aligned}
\end{equation}
\begin{equation}
\begin{aligned}
    y_2 &= l_1\cos(x_2) + l_2\cos(x_3-x_2) \\
    &+ l_3\cos(x_4-x_2-x_3)
\end{aligned}
\end{equation}
where $l_1=0.5$, $l_2=0.5$ and $l_3=1$ in this case. The parameters $\mathbf{x}$ have a Gaussian prior $\mathbf{x} \sim p(\mathbf{x}) = \mathcal{N}(0, \sigma^2 \cdot \mathbf{I})$ with $\sigma^2 = [\frac{1}{16}, \frac{1}{4}, \frac{1}{4},\frac{1}{4}]$. The inverse problem is to obtain all possible solutions in the $\mathbf{x}$-space given any observed 2D positions $\mathbf{y}^*$. 

\subsection{Ballistics task}
The forward process in this example can be interpreted as an object being thrown from position $(x_1, x_2)$ with angle $x_3$ and initial velocity $x_4$ and the output is the object's final position on the horizontal axis $y = T_1(t^*)$ where $t^*$ is the solution of $T_2(t^*)=0$. The object's trajectory $\mathbf{T}(t)$ can be computed analytically as follows:
\begin{equation}
    T_1(t) =x_1 - \frac{v_1 m}{k}\cdot \left( \exp\left(-\frac{kt}{m}\right)-1\right)
\end{equation}
\begin{equation}
\begin{aligned}
    T_2(t) &=x_2 - \frac{m}{k^2} \cdot ( \left( gm + v_2 k \right) \\ &\cdot \left(\exp\left(-\frac{kt}{m} \right)-1 \right) + gtk )
\end{aligned}
\end{equation}

Given gravity $g$, object mass $m$ and air resistance $k$. $v_1=x_4 \cdot \cos(x_3)$ and $v_2 = x_4 \cdot \sin(x_3)$ are the horizontal and vertical velocity respectively. The prior distributions are defined as $x_1 \sim \mathcal{N}(0, \frac{1}{4})$, $x_2 \sim \mathcal{N}(\frac{3}{2}, \frac{1}{4})$, $x_3 \sim \mathcal{U}(9^{\circ},72^{\circ})$ and $x_4 \sim \textup{Poisson}(15)$.

\section{Real-World Applications: Additional Details}

\subsection{Crystal design problem}
\paragraph{Background}
We apply our approach to a challenging real-world application in materials design, specifically for modeling the electronic properties of complex metal oxides. Metal oxides can exhibit significant changes in their electronic and magnetic properties in the presence of external perturbations such as strain, and electric and magnetic fields, with major implications for the design of neuromorphic and quantum devices \cite{RN6466, RN6431, RN6467}.

\paragraph{Methods for Crystal Prediction Experiment}
Lattice constants and angles $a$, $b$, $c$, $\alpha$, $\beta$, $\gamma$ were sampled uniformly within a range of 10\% deviation from the equilibrium crystal parameters: $a=b=c=3.914$ \AA ~and  $\alpha=\beta=\gamma=90^{\circ}$. Within the perturbed ranges (± 10\% of equilibrium value) in the lattice constants of the training data, the band gaps of the crystals were found to vary between 0 to 2.8 eV, representing a very wide range in energies. For reference, the band gap of the unperturbed crystal computed at this level of theory is 2.37 eV. In our experiment, we select an arbitrary target of 0.5 eV to generate our structures and compare the performance of our model with the existing ones. 

\begin{figure}[h!]
    \centering
    \includegraphics[width=0.4\textwidth]{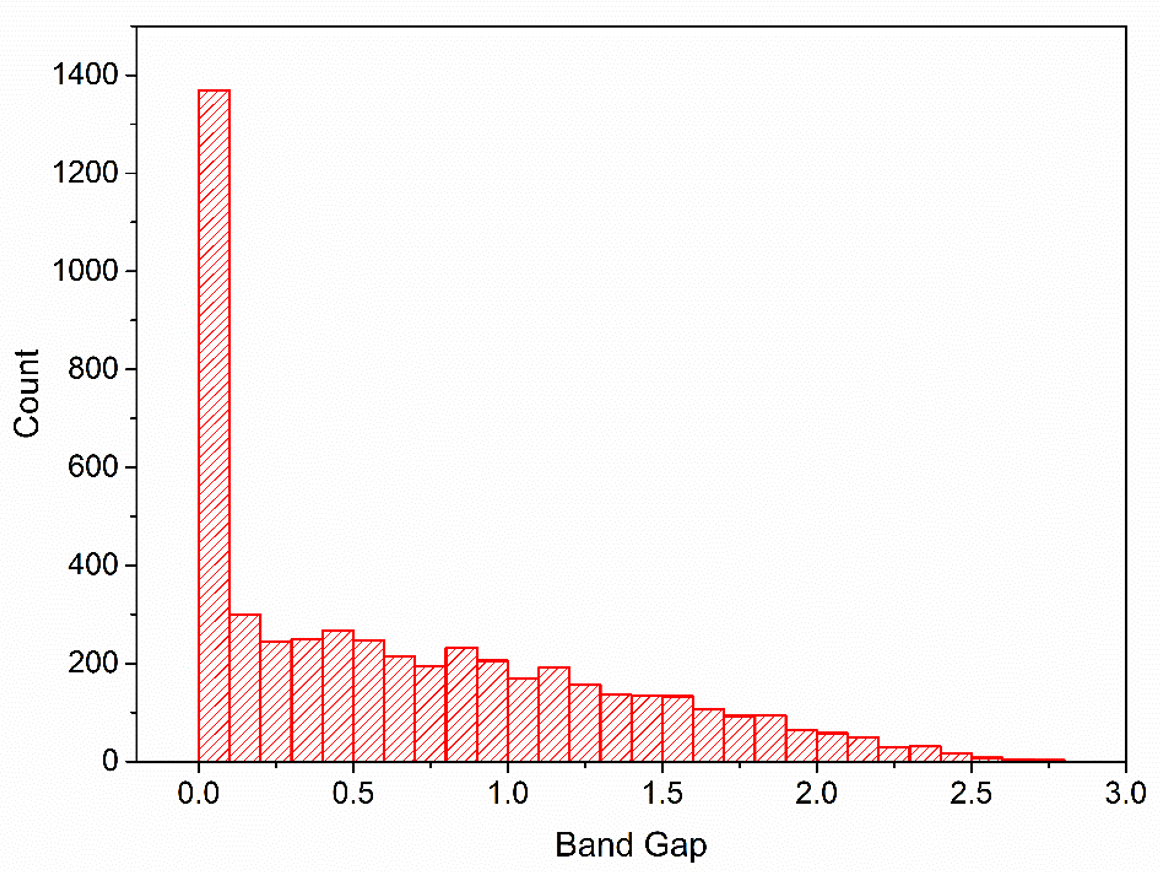}
    \caption{Distribution of calculated band gaps in the dataset.}
    \label{fig:crystal2}
\end{figure}

A total of 5000 structures were generated and band gaps were obtained using density functional theory (DFT). The distribution of calculated band gaps in the dataset is shown in Fig. \ref{fig:crystal2}. The DFT calculations were performed with the Vienna Ab Initio Simulation Package (VASP) \cite{RN138, RN144}. The Perdew-Burke-Ernzerhof (PBE)\cite{RN145} functional within the generalized-gradient approximation (GGA) was used for electron exchange and correlation energies. The projector-augmented wave method was used to describe the electron-core interaction \cite{RN141, RN138}. The on-site Coulomb interaction was included using the DFT+U method by Dudarev, et al.\cite{RN147} in VASP using a Hubbard parameter U = 4 eV for the Ti based on previous studies \cite{RN6481, RN6480}.  A kinetic energy cutoff of 500 eV was used. All calculations were performed with spin polarization. The Brillouin zone was sampled using a Monkhorst-Pack scheme with an $8\times8\times8$ grid \cite{RN148}. 

\begin{table*}[!h] 
\centering
\caption{Generated lattice parameters for the crystal band gap experiment for $y^*=0.5$ eV} 
\label{tab:crystal}
\begin{tabular}{@{}ccccccc@{}}
\toprule
Method    & a     & b     & c     & $\alpha$       & $\beta$       & $\gamma$       \\ \midrule
iPage 1   & 4.412 & 3.524 & 3.923 & 94.132  & 95.124  & 106.975 \\
iPage 2   & 3.621 & 4.125 & 4.692 & 85.133  & 102.299 & 77.432  \\
INN   & 4.134 & 3.931 & 4.016 & 92.728  & 89.834  & 92.207  \\
cINN  & 3.972 & 3.952 & 3.923 & 93.357  & 89.815  & 90.199  \\
cVAE  & 4.073 & 4.133 & 4.126 & 90.152  & 89.128  & 91.298  \\
NA    & 3.510 & 4.602 & 4.825 & 94.988  & 95.016  & 103.775 \\ \bottomrule
\end{tabular}
\end{table*}

\begin{table*}[!h]  
\centering
\caption{Hyperparameter settings for each benchmark task} 
\label{tab:setting}
\begin{tabular}{@{}cccccc@{}}
\toprule
Parameters                              & Sinewave  & Robotic Arm & Ballistics  & Crystal  & Architecture  \\ \midrule
Number of invertible blocks             & 4         & 6           & 6           & 4  & 6       \\
Number of fully connected (fc)   layers & 3         & 3           & 3           & 3  & 3       \\
Number of neurons in each fc            & 128       & 256         & 256         & 64  & 512       \\
Activation function in each fc          & ReLU      & Leakly ReLU & Leakly ReLU & ReLU  & ReLU    \\
Optimizer for invertible training                               & Adam      & Adam        & Adam        & Adam & Adam      \\
Optimizer for localization                               & Adam      & Adam        & Adam        & Adam & Adam     \\
Batch size                              & 1024      & 512         & 512         & 256  & 1024      \\
Number of training epoch                & 1000      & 500         & 500         & 1000 & 10000     \\
Learning rate (decay) for invertible training                   & 1e-3-1e-5 & 1e-2-1e-4   & 1e-2-1e-4   & 1e-3-1e-5 & 1e-3-1e-5 \\
Learning rate for localization                   & 1e-3 & 5e-3   & 5e-3   & 1e-3 & 1e-4 \\ \bottomrule
\end{tabular}
\end{table*}

\subsection{Architected materials design problem}
\paragraph{Background}

Architected materials on length scales from nanometers to meters are widely used fro many practical applications \cite{mao2020designing}. Controlling material architecture –a complex interplay of topology, material distribution, and constituent material behavior– is a powerful way to create materials with tailored and often unprecedented properties. We, therefore, envision a shift towards a materials-by-design approach that takes advantage of rapid advancements in material synthesis, and in particular additive manufacturing, to produce novel material systems across many length scales and with multiple functionalities. Examples of interest include mechanical, seismic, and acoustic meta-materials, additively manufactured lattices and foams, nanostructured materials, optimized architectures, and design algorithms, materials for extreme environments, and natural architected materials. 

\paragraph{Data generation}
We generate the data from topology optimization approaches \cite{bendsoe2013topology}. Based on the symmetric properties, topology optimization is an effective way to obtain the desired architecture given a specific elastic property, e.g., Young's Modulus $E$. The input is the pixel matrix for the element of an architected material, and the output is the effective mean Young's modulus of the corresponding architected materials. The size of the dataset is one million configurations, including different groups with various porosities, typically ranging from 0.25 to 0.75.

\section{Details of Hyperparameter Settings}
In Table \ref{tab:setting}, we provide a summary of the hyperparameter settings, including the neural network model, number of training epochs, batch size, invertible architectures, optimizer, learning rate decay and etc, for each benchmark task. 

\end{document}